\documentclass[10pt,twocolumn,letterpaper]{article}

\usepackage{cvpr}              

\usepackage{pifont}      
\usepackage{colortbl}    
\usepackage{pgfplots}    
\usepackage{arydshln}
\pgfplotsset{compat=1.17}
\usepgfplotslibrary{groupplots}
\usetikzlibrary{calc,arrows.meta,shapes.geometric}

\definecolor{oursgray}{HTML}{E2EFDA}
\usepackage{algorithm}
\usepackage{algpseudocode}
\algnewcommand{\IIf}[2]{\State \algorithmicif\ #1\ \algorithmicthen\ #2}
\definecolor{evogreen}{HTML}{70AD47}
\definecolor{evocolor}{HTML}{7E57C2}
\definecolor{autocolor}{HTML}{E69F00}
\definecolor{mintcolor}{HTML}{2A9D78}
\definecolor{wf0}{HTML}{0D0887}\definecolor{wf1}{HTML}{6A00A8}\definecolor{wf2}{HTML}{B12A90}
\definecolor{wf3}{HTML}{E16462}\definecolor{wf4}{HTML}{F28F2C}\definecolor{wf5}{HTML}{D4B000}

\definecolor{cvprblue}{rgb}{0.21,0.49,0.74}
\usepackage[pagebackref,breaklinks,colorlinks,allcolors=cvprblue]{hyperref}
\makeatletter
\providecommand{\theHALG@line}{}
\renewcommand{\theHALG@line}{\thealgorithm.\arabic{ALG@line}}
\makeatother
\def\confName{CVPR}
\def\confYear{2027}

\title{\texorpdfstring{\raisebox{-0.3\height}{\includegraphics[height=2.5em]{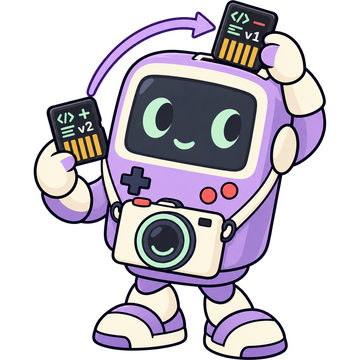}}\hspace{0.4em}\begin{tabular}[c]{@{}c@{}}EvoViM: Recursively Self-Evolving Visual Agents\\for Exploration with Deferred Queries\end{tabular}}{EvoViM: Recursively Self-Evolving Visual Agents for Exploration with Deferred Queries}}

\author{%
Xinyu Cao$^{1,2}$\thanks{Xinyu Cao conducted this research during an internship at Shanghai Jiao Tong University.}
\quad Tianjie Ju$^{3}$
\quad Zhuosheng Zhang$^{1}$\thanks{Corresponding author.}\\
{\small $^{1}$Shanghai Jiao Tong University \quad
$^{2}$Hunan University \quad
$^{3}$National University of Singapore}\\
{\tt\small lewisxinyu@hnu.edu.cn \quad jutianjie@nus.edu.sg \quad zhangzs@sjtu.edu.cn}
}

\begin{document}
\maketitle
\raggedbottom

\begin{figure*}[t]
  \centering
  \includegraphics[width=\linewidth]{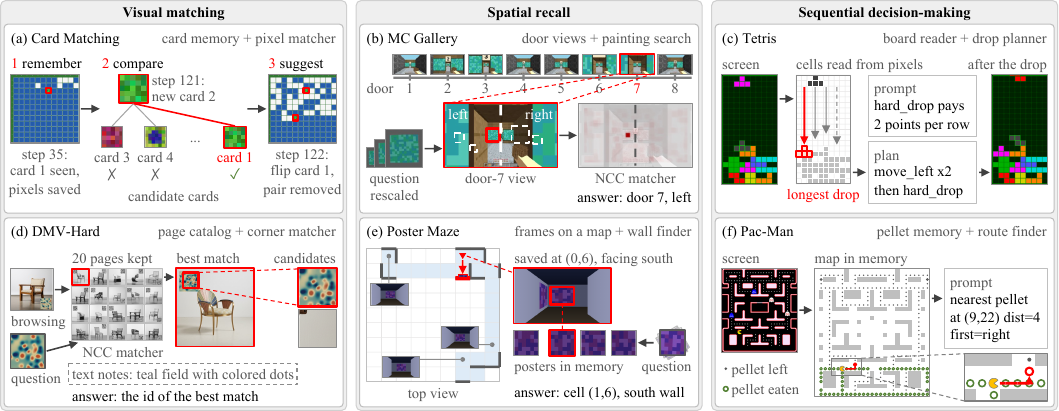}
  \caption{EvoViM evolves task-specific visual programs from one generic starting program. Each panel shows an evolved program on a held-out episode, with the evolved mechanisms named in its header. Red marks the evidence the program finds or the action it suggests.}
  \label{fig:teaser}
\end{figure*}

\begin{abstract}

Multimodal large language model (MLLM) agents have shown promising capabilities in task-oriented environments. 
However, they still struggle when queries are deferred until after exploration, requiring them to retain visual evidence before knowing which details will matter.
We introduce EvoViM, a recursively self-evolving framework in which a frozen multimodal model turns its own interaction experience into one visual program for memory and skills.
The program retains images and computes directly on pixels, and its operations decide how this evidence is used in each task.
During memory evolution, the same model traces failures in its visual trajectories to the operations that lost or misused evidence and revises them.
It also rewrites the notes and analysis tools that guide later revisions, which then build on earlier findings and skip changes that have already failed.
We evaluate EvoViM with three multimodal models on six tasks covering visual matching, spatial recall, and sequential decision-making, including three tasks with deferred queries. 
Across the six tasks, EvoViM improves the average score over AutoMem, the strongest baseline in the main comparison, by 31.3--47.6 points on held-out episodes. 
Programs written once from the initial environment instructions perform substantially worse, highlighting the value of interaction experience in shaping how visual evidence is retained and used.
Code and dataset are available at \url{https://evovim.github.io/}.
\end{abstract}

\section{Introduction}
\label{sec:intro}

Multimodal large language model (MLLM) agents have made rapid progress in task-oriented environments, where they navigate websites~\citep{zheng2024seeact}, operate computers~\citep{xie2024osworld}, and play video games~\citep{Cradle,GameWorld} from screenshots.
In these environments, the instruction is given before interaction begins and guides what the agent looks for and remembers.
However, some tasks require an agent to explore before it receives the question~\citep{openeqa}.
For example, an agent may walk through a gallery and later be asked where a re-photographed painting hung, although its instructions never mentioned the paintings (Figure~\ref{fig:teaser}(b)).
We study exploration with deferred queries, in which the queries are revealed only after the agent has explored the environment.

Existing agents keep past observations in memories that are either designed by hand or evolved from experience.
Hand-designed memories store text notes or frames~\citep{JARVIS-1,MrSteve,han2026vista}, and their designers fix how the stored evidence is retrieved and used.
Even when such memories store every frame, they adapt poorly to different tasks.
Memory-evolution methods learn the rules from experience and adapt them to the task~\citep{wu2026automemautomatedlearningmemory,xiong2026learningcontinuallylearnmetalearning}, but the memory they evolve is text, which cannot record the visual details that separate near-identical images~\citep{memeye,dmvbench}.
These methods therefore break down when the decisive evidence is hard to put into words.

In this paper, we study exploration with deferred queries in multimodal settings and propose EvoViM, a recursively self-evolving framework that learns memory and skills from interaction as one visual program.
We treat memory and skills as two uses of experience.
The program holds the observations of the current episode, and its operations, shaped by earlier episodes, decide what to keep and how to use it.
These operations process pixels directly to retrieve evidence from past observations or to suggest actions from the current state.
In EvoViM, one frozen MLLM executes tasks with the current program and revises the program between generations.
During revision, it reads the visual trajectories of development episodes, traces each failure to the operation that lost or misused the evidence, and rewrites that operation.
It also rewrites the notes and analysis scripts that guide later revisions, so that it can build on earlier findings and avoid repeating failed changes.

We first construct three interactive visual tasks with deferred queries.
The agent must retain visual details that are difficult to describe in words, then use them to answer questions revealed after exploration.
Our comparisons show that evolved text memory loses these visual distinctions, while hand-designed visual memory often fails to turn retrieved evidence into correct answers.
Programs written once from the initial environment instructions alone often prepare for the wrong questions.

We then evaluate EvoViM with three MLLMs on six tasks that test visual matching, spatial recall, and sequential decision-making, including the three tasks with deferred queries.
EvoViM achieves the highest average score with every model and exceeds the strongest baseline by 31.3 to 47.6 points.
These results show that MLLMs can evolve visual memory programs from interaction experience to preserve and use visual information that is difficult to describe in words, supporting exploration with deferred queries.

Our contributions are as follows.
\begin{itemize}
  \setlength{\itemsep}{0pt}
  \item We study exploration with deferred queries and build three multimodal environments on published benchmarks.
  \item We propose EvoViM, in which a frozen MLLM evolves one visual program for memory and skills and revises the guidance for its own later revisions.
  \item We show on six tasks with three models that evolution from a common starting program yields operations suited to each task, and we trace how failures during interaction shape each revision.
\end{itemize}

\begin{figure*}[t]
  \centering
  \includegraphics[width=\linewidth]{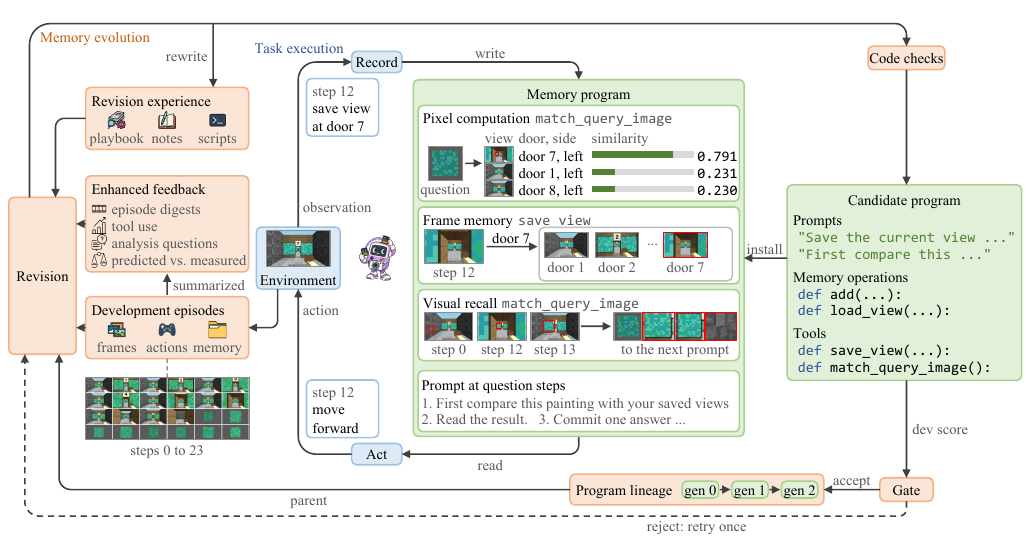}
  \caption{Overview of EvoViM. In task execution (blue), the agent records each outcome and consults the memory program (green) before acting. In memory evolution (orange), the same frozen MLLM revises the program from development episodes. A candidate is kept only if it passes code checks and the gate, and the model also rewrites the revision experience that guides its later revisions.}
  \label{fig:framework}
\end{figure*}
\section{Related Work}
\label{sec:related}

\subsection{Multimodal Agents}
Multimodal agents use visual observations to act in games, websites, and desktop environments.
Cradle uses screenshots with keyboard and mouse control to play games and operate desktop software~\citep{Cradle}.
SeeAct combines screenshots with HTML elements to ground actions on live websites~\citep{zheng2024seeact}, and OSWorld evaluates agents on real computer tasks, including workflows that span several applications~\citep{xie2024osworld}.
Game benchmarks such as BALROG, VideoGameBench, and GameWorld show that long-horizon play remains difficult~\citep{BALROG,VideoGameBench,GameWorld}.
Some decisions require evidence from observations that have already left the screen~\citep{rngbench}.
OpenEQA directly evaluates the use of past observations through questions about previously explored physical environments~\citep{openeqa}.
These works do not address how agents can improve their visual memory from experience gained during interaction.

\subsection{Agent Memory and Skills}
To reuse past observations, many agents keep them as textual records or summaries~\citep{park2023generative,packer2023memgpt,wu2025resum}, which can omit visual details needed for later tasks~\citep{memeye,dmvbench}.
Visual memories retain images for retrieving past experience~\citep{JARVIS-1} or recalling explored places~\citep{MrSteve,yang20253dmem}, while VISTA archives every frame~\citep{han2026vista}.
Experience can also be reused through learned operations.
Voyager accumulates executable action skills~\citep{Voyager}, and MemSkill evolves skills for building and updating textual memory~\citep{memskill2026}.
AtlasVA connects visual memory to skill learning by evolving spatial maps from trajectory statistics and using them to guide reinforcement learning~\citep{AtlasVA}.
However, the rules for organizing and using visual evidence in these systems are still predefined.

\subsection{Recursive Self-Improvement}
Self-improving agents learn from their own experience either by updating model weights~\citep{selfrewarding2024,absolutezero2025,mstar2024,agent0vl2025,sima22025} or by revising the code around a frozen model, such as its prompts and tools~\citep{codewithimage2026,physevo2026}.
Recursive methods also improve the procedure that makes these revisions~\citep{zhou2026hierarchicalselfimprovementframeworktaskspecific}.
STOP improves the program that improves code~\citep{zelikman2024stop}, Gödel Agent rewrites its own logic~\citep{yin2025godel}, and the Darwin Gödel Machine keeps an archive of coding agents that modify their own code~\citep{zhang2026dgm}.
ALMA, AutoMem, and MemPro search over memory programs~\citep{xiong2026learningcontinuallylearnmetalearning,wu2026automemautomatedlearningmemory,mempro2026}, but the memory and the feedback that guides its revision stay in text.
MemEvo evolves memory mechanisms for streaming video, whose frames do not depend on the actions of the model~\citep{memevo2026}.
AutoMMemo designs memory programs for multimodal experience, which are queried only once before a task begins~\citep{sui2026learninglearnmultimodalexperience}.
However, these methods do not study how agents can learn to retain and use visual evidence in exploration with deferred queries.
\section{Why Do Existing Memory Designs Fall Short with Deferred Queries?}
\label{sec:pilot}

Current MLLM agents can follow instructions to complete tasks in visual environments.
With deferred queries, however, the agent explores with only basic environment instructions and receives the questions afterward.
It must decide what to retain without knowing which visual details will matter or how those details will be used.

We build three tasks to examine how existing memory designs handle this setting.
In MC Gallery and Poster Maze, the agent must recall where a re-photographed painting or poster hung, or report that it was never seen.
In DMV-Hard, it must identify a previously visited product page from a transformed crop of a photo detail or a generated pattern.
We compare evolved text memory with a hand-designed visual memory, then test whether the model can infer the required operations from the initial instructions alone.
We find three key limitations that cause existing memory designs to fail in exploration with deferred queries.

\begin{table}[t]
  \centering
  \footnotesize
  \setlength{\tabcolsep}{3pt}
  \caption{Scores of different memory designs on the three tasks with deferred queries.}
  \label{tab:fixed}
  \begin{tabular}{@{}lcccccc@{}}
    \toprule
    & \multicolumn{3}{c}{Claude Haiku 5.5} & \multicolumn{3}{c}{Qwen3.8-27B} \\
    \cmidrule(lr){2-4}\cmidrule(l){5-7}
    Method & MC & Maze & DMV & MC & Maze & DMV \\
    \midrule
    Basic image tools & 15.0 & 18.8 & 13.8 & 23.8 & 23.8 & 7.9 \\
    Evolved text memory~\citep{wu2026automemautomatedlearningmemory} & 23.8 & 16.3 & 42.1 & 25.0 & 17.5 & 37.9 \\
    VISTA~\citep{han2026vista} & 20.0 & 16.3 & 22.1 & 25.0 & 21.3 & 20.4 \\
    \quad + matcher & 18.8 & 16.3 & 80.8 & 47.5 & 15.0 & 81.7 \\
    One-shot program & 13.8 & 20.0 & 28.8 & 13.8 & 25.0 & 14.2 \\
    \bottomrule
  \end{tabular}
\end{table}

\paragraph{Text descriptions lose visual distinctions.}
Text memory evolved from experience~\citep{wu2026automemautomatedlearningmemory} answers most DMV-Hard questions about photo details but almost none about generated patterns (Figure~\ref{fig:dmv_kind}).
Its descriptions of one pattern often fit several others, so the stored notes cannot distinguish the queried item.
On MC Gallery and Poster Maze, its scores are no higher than those obtained by answering \emph{not seen} to every question (Table~\ref{tab:fixed}).

\paragraph{A visual match does not ensure a correct answer.}
To preserve these distinctions, we test a hand-designed visual memory based on VISTA~\citep{han2026vista} that keeps every frame for later inspection.
This archive alone still scores below evolved text memory on DMV-Hard.
Adding a matcher that runs at every step lets it answer most DMV-Hard questions, as the matched page identifies the product.
In MC Gallery and Poster Maze, the matcher usually retrieves the right view for previously seen pictures, but the agent often fails to answer correctly.
The program lacks operations that associate the match with a location and determine whether the queried picture was seen at all.

\paragraph{One-shot programs often prepare for the wrong questions.}
We therefore test whether the model can infer the required operations from the initial environment instructions alone.
One gallery program records the order of the doors passed, whereas the later questions ask where each painting hung.
On MC Gallery and Poster Maze, these programs show no clear gain over the initial set of basic image tools.
The questions encountered during interaction provide the experience needed to revise what the program keeps and how it uses that evidence.
\section{Method}
\label{sec:method}

EvoViM turns the experience of an agent into a visual program for memory and skills (Figure~\ref{fig:framework}).
Within an episode, the program stores observations and maintains state.
Its operations, which decide what to store and how to use it, are shaped by earlier episodes.
One frozen MLLM executes tasks with the program and, between generations, revises these operations from development episodes.
It also revises the guidance and analysis tools used to diagnose failures in later generations, so the evolution is recursive.

\subsection{Memory and Skills in One Visual Program}
\label{sec:method_inner}

The visual program defines memory operations and image-processing tools in Python, with prompts that guide the model in using them.
The code and prompts remain fixed during an episode, while stored observations and state are updated through interaction.
These memory contents are reset at the start of the next episode.
In MC Gallery (Figure~\ref{fig:framework}), for example, the evolved program instructs the model to save a view at each door during the walk.
When a query painting appears, a matching tool compares it with the saved views and reports that the best match lies to the left of door 7.
The model then submits this answer.

The starting program stores lossless frames with text metadata and has no pixel matcher.
Evolution can change what frame memory retains and how it is organized, allowing the program to store whole views, object crops, or pixel features linked to locations.
Pixel computation lets it process this stored evidence alongside the current image.
These operations preserve the visual distinctions between similar images that text notes lose (Section~\ref{sec:pilot}).

To make stored evidence available at a decision, visual recall places selected frames, crops, or montages in the next model request.
Revisions can change which text and images appear in each prompt, so the program determines what the model sees when it uses memory.
The same code can read the current frame and the accumulated state to score candidate actions or suggest the next action.
In Card Matching, for example, the program combines remembered card locations with pixel matching to suggest the next flip.

During task execution, the model uses these operations in a \emph{Record} phase followed by an \emph{Act} phase (Algorithm~\ref{alg:inner}).
It first examines the observed outcome of its previous action and decides what to record, then consults memory to choose the next action.
Each phase starts a new conversation, while the episode memory remains available through the program.
Within that conversation, every request includes the current image, and tool results enter the next request.
The model then submits its chosen action through the fixed interface.
The model and this execution procedure remain fixed during evolution.

\begin{algorithm}[!t]
  \caption{Task execution with a visual memory program.}
  \label{alg:inner}
  \small
  \begin{algorithmic}[1]
    \Require program $P$, frozen model $M$, task seed
    \Require episode steps $T$, request cap $K = 20$
    \State $S \gets \emptyset$ \Comment{episode memory}
    \For{$t = 0, \dots, T-1$}
      \State $o_t \gets$ current observation
      \If{$t > 0$}
        \State \Call{Phase}{\textsc{Record}, $o_t$}
      \EndIf
      \State $a_t \gets \Call{Phase}{\textsc{Act}, o_t}$
      \If{$a_t$ is missing or illegal}
        \State $a_t \gets$ fallback action
      \EndIf
      \State the host executes $a_t$
    \EndFor
    \Statex
    \Function{Phase}{$\phi, o_t$}
      \State $H \gets \emptyset$ \Comment{phase conversation}
      \For{$k = 1, \dots, K$}
        \State $y \gets M(P.\Call{Prompt}{\phi, S}, o_t, H)$
        \If{$y$ is a memory command}
          \State $(S, r) \gets P.\Call{Execute}{y, S, o_t}$
          \State $H \gets \Call{UpdateContext}{H, y, r}$
        \ElsIf{$\phi = \mathrm{Act}$ and $y$ commits $a$}
          \State \Return $a$
        \ElsIf{$y$ ends the phase}
          \State \Return $\varnothing$
        \EndIf
      \EndFor
      \State \Return $\varnothing$
    \EndFunction
  \end{algorithmic}
\end{algorithm}

\subsection{Revising Programs from Interaction Feedback}
\label{sec:method_outer}

The development episodes recorded during task execution provide the evidence for revising the program.
We first run the starting program to obtain the episodes used by the first revision.
At every generation, the model uses a command-line coding tool to inspect the parent program and its development episodes, including frames labeled by step, memory snapshots, and tool results.
Records of earlier candidates show how previous changes affected task outcomes on the development episodes.

To help the program generalize to new episodes, each revision uses evidence from two development seeds.
The model is asked to check each proposed change against both episodes and explain why it should work on unseen instances.
The revision instructions prohibit embedding facts from development episodes in the code and require the program to obtain episode-specific information from observations during execution.
Feedback reports results separately for each seed, so later revisions can identify changes that help one episode at the expense of another.

Enhanced feedback guides revision by connecting task outcomes to how the program handled visual evidence.
Episode digests summarize the outcomes, while tool-use statistics show which memory operations the agent called.
The revision instructions include fixed analysis questions that help the model identify where visual evidence was lost or misused.
Based on this diagnosis, the model must propose a structural revision, such as a new storage schema or a tool that processes visual evidence.
Before evaluation, it predicts how the candidate will change the score and tool use.
Later feedback compares these predictions with the measured changes, so the model can check whether the revision had the intended effect.

The model preserves what it learns from these revisions in a playbook, notes, and analysis scripts.
The notes record findings from earlier attempts and guide revisions to the playbook used for later diagnosis and editing.
Analysis scripts automate checks on recorded episodes and tests of proposed changes, allowing later revisions to reuse or extend earlier tools.
These files pass to every later revision, including after a candidate is rejected.
The selection rule, evolution budget, and access to evidence remain fixed as this revision experience accumulates.

\subsection{Evaluating and Retaining Candidates}
\label{sec:method_gate}

Before running development episodes, the host checks the candidate by replaying recorded observations with a stub model that calls every registered tool.
This tests the tool calls and checks that saved episode state can be restored without making model requests.
A candidate that passes these checks then runs on the same two development seeds as its parent, with one episode per seed.

The gate accepts a candidate if its mean development score is higher.
For a score tie, the candidate must succeed on every development seed and use fewer steps or tokens without increasing the other.
If the candidate fails the code checks or the gate, the model revises the same parent once more using the rejected code and failure evidence.

After the final generation, the last accepted program is fixed and evaluated on held-out seeds disjoint from the development seeds.
The agent starts each held-out episode with empty memory and accumulates observations and state using the evolved program.

Algorithm~\ref{alg:outer} summarizes the procedure, with candidate records $R$ and revision experience $X$ retained across attempts.
The records include predicted changes $\pi$, code-check outputs $L$, and the measured outcomes.
The model updates $X$ during revision, while the gate determines whether the candidate replaces the parent program.

\begin{algorithm}[!t]
  \caption{Evolution of one program lineage.}
  \label{alg:outer}
  \small
  \begin{algorithmic}[1]
    \Require initial program $P_0$, frozen model $M$
    \Require development seeds $\mathcal{D}$, held-out seeds $\mathcal{H}$
    \Require generations $G$, initial revision experience $X_0$
    \State $P \gets P_0$, \quad $X \gets X_0$
    \State $E \gets \Call{Play}{M, P, \mathcal{D}}$
    \State $R \gets \emptyset$ \Comment{candidate records}
    \For{$g = 1, \dots, G$}
      \For{$k = 1, 2$} \Comment{one candidate, then one retry}
        \State $F \gets \Call{Feedback}{E, R}$
        \State $(C, \pi, X) \gets \Call{Revise}{M, P, E, F, R, X}$
        \State $(\mathit{ok}, L) \gets \Call{Checks}{C}$
        \State $E_C \gets \emptyset$
        \If{$\mathit{ok}$}
          \State $E_C \gets \Call{Play}{M, C, \mathcal{D}}$
          \State $\mathit{ok} \gets \Call{Gate}{E_C, E}$
        \EndIf
        \State $R \gets R \cup \{(C, \pi, E_C, \mathit{ok}, L)\}$
        \If{$\mathit{ok}$}
          \State $P \gets C$, \quad $E \gets E_C$
          \State \textbf{break}
        \EndIf
      \EndFor
    \EndFor
    \State \Return \textproc{Score}$\bigl(\textproc{Play}(M, P, \mathcal{H})\bigr)$
  \end{algorithmic}
\end{algorithm}
\section{Experiments}
\label{sec:exp}

\subsection{Experimental Setup}
\label{sec:exp_setup}

\begin{table}[!t]
  \centering
  \footnotesize
  \setlength{\tabcolsep}{3pt}
  \caption{Task settings and held-out evaluation budgets.}
  \label{tab:envs}
  \begin{tabular}{@{}lllrr@{}}
    \toprule
    Task & To recall & Goal & Steps & Episodes \\
    \midrule
    Card Matching & 144 cards & 72 pairs & 360 & 5 \\
    DMV-Hard & 100 product pages & 24 answers & 124 & 10 \\
    MC Gallery & 16 paintings ($12{\times}12$) & 8 answers & 60 & 10 \\
    Poster Maze & 16 posters ($6{\times}6$) & $\leq 8$ answers & 140 & 10 \\
    Tetris & -- & task progress & 100 & 5 \\
    Pac-Man & -- & task progress & 100 & 5 \\
    \bottomrule
  \end{tabular}
\end{table}

\begin{figure*}[!t]
  \centering
  \includegraphics[width=\linewidth]{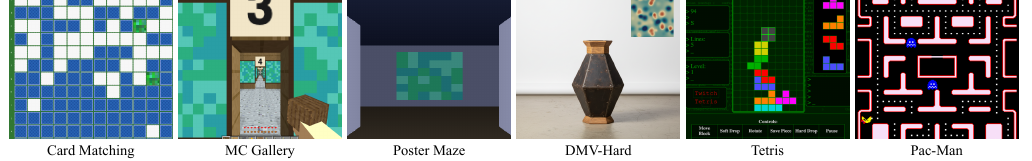}
  \caption{Example observations from the six evaluation tasks.}
  \label{fig:envs}
\end{figure*}

\paragraph{Tasks.}
We evaluate EvoViM on six tasks that test visual matching, spatial recall, and sequential decision-making (Table~\ref{tab:envs} and Figure~\ref{fig:envs}).
For visual matching, Card Matching uses noise-texture cards on a $12{\times}12$ board from RNG-Bench~\citep{rngbench}.
Unmatched cards turn face down again, so the agent must remember their appearance and location to find pairs.
DMV-Hard adapts the product-browsing task of DMV-Bench~\citep{dmvbench}.
We add generated patterns to the product photos, then ask the agent to identify products from earlier browsing sessions using transformed crops of photo details or patterns.

For spatial recall, we build MC Gallery in GameWorld~\citep{GameWorld} and Poster Maze in RNG-Bench.
The agent explores the environment and is later shown a re-photographed picture and asked where it hung, with unseen pictures as lures.
These two tasks and DMV-Hard reveal their queries only after exploration.

Tetris and Pac-Man test sequential decision-making in GameWorld.
They show their full state and test whether evolved programs support decisions from the current observation.
Pac-Man uses the same starting layout across episodes.

\paragraph{Metrics.}
We report the fraction of pairs cleared for Card Matching and answer accuracy for DMV-Hard.
MC Gallery asks about six seen paintings and two unseen lures.
Poster Maze asks about at most six clearly observed posters and two lures, so the number of questions depends on exploration.
Both tasks divide the number of correct answers by eight, and a lure counts as correct only when answered \emph{not seen}.
Always giving this answer scores 25\%.

For Tetris and Pac-Man, we use GameWorld's normalized task progress.
Tetris awards points for both dropping pieces and clearing lines.

Scores are reported as percentages, with equal weight given to each task in the overall average.
We compute 95\% confidence intervals using 10,000 paired bootstrap resamples of the held-out seeds within each task.

\paragraph{Baselines.}
We choose baselines by how they retain past observations and whether their memory operations evolve.
The No memory baseline provides a reference without stored observations, while Rolling summary compresses past interactions into text in the style of ReSum~\citep{wu2025resum}.
Hand-crafted text provides a note-taking tool and is the starting program for AutoMem.
Hand-crafted visual provides five tools for frame storage, annotation, metadata search, recall, and deletion.
EvoViM starts from this program.
These four baselines keep their memory mechanisms fixed.
AutoMem~\citep{wu2026automemautomatedlearningmemory} evolves a text memory program from interaction.
To evaluate AutoMem in visual environments, we provide the current image at each execution request while retaining its original text memory design.
For this comparison, we use the same MLLM to execute and revise its program.
AutoMem accepts a candidate only when its mean development score improves.

\paragraph{Models and protocol.}
We use GPT-6.1 Sol~\citep{openai2026gpt61sol}, Claude Haiku 5.5~\citep{anthropic2026haiku55}, and the open-weight Qwen3.8-27B~\citep{qwen2026qwen38}.
Claude Haiku 5.5 revises its programs in Claude Code~\citep{anthropic2026claudecode}, while GPT-6.1 Sol and Qwen3.8-27B use Codex~\citep{openai2026codex}.
Both evolving methods develop a separate program for each task and model over three generations on the same two development seeds.
Each generation allows one candidate and at most one retry.

\subsection{Main Results}
\label{sec:exp_main}

\paragraph{Overall performance.}
EvoViM exceeds AutoMem, the strongest baseline in the main comparison, by 31.3 to 47.6 points in average score (Table~\ref{tab:main}).
The paired bootstrap confidence intervals for this improvement exclude zero with all three models.
When scores are averaged within each ability group, EvoViM also ranks first, except for sequential decision-making with GPT-6.1 Sol.

\begin{table*}[!t]
  \centering
  \small
  \setlength{\tabcolsep}{4pt}
  \caption{Mean scores (\%) with per-task standard deviations across seeds. \textbf{Bold} and \underline{underline} mark the best and second-best means within each model.}
  \label{tab:main}
  \begin{tabular}{@{}lccccccc@{}}
    \toprule
    & \multicolumn{2}{c}{Visual matching} & \multicolumn{2}{c}{Spatial recall} & \multicolumn{2}{c}{Sequential decision-making} & \\
    \cmidrule(lr){2-3}\cmidrule(lr){4-5}\cmidrule(lr){6-7}
    Method & Card Matching & DMV-Hard & MC Gallery & Poster Maze & Tetris & Pac-Man & Avg. \\
    \midrule
    \multicolumn{8}{@{}l}{\textbf{GPT-6.1 Sol}} \\
    \hspace{1em}No memory & 0.3{\scriptsize$\pm$0.6} & 0.0{\scriptsize$\pm$0.0} & \underline{25.0}{\scriptsize$\pm$0.0} & 25.0{\scriptsize$\pm$0.0} & 34.7{\scriptsize$\pm$9.1} & \textbf{45.3}{\scriptsize$\pm$21.5} & 21.7 \\
    \hspace{1em}Rolling summary & 8.6{\scriptsize$\pm$4.5} & 1.7{\scriptsize$\pm$2.2} & 11.3{\scriptsize$\pm$12.4} & 17.5{\scriptsize$\pm$14.7} & \textbf{51.1}{\scriptsize$\pm$19.6} & 42.6{\scriptsize$\pm$14.6} & 22.1 \\
    \hspace{1em}Hand-crafted text & 0.8{\scriptsize$\pm$0.8} & 11.3{\scriptsize$\pm$6.8} & \underline{25.0}{\scriptsize$\pm$0.0} & 17.5{\scriptsize$\pm$8.7} & 36.9{\scriptsize$\pm$4.6} & \underline{44.5}{\scriptsize$\pm$20.3} & 22.7 \\
    \hspace{1em}Hand-crafted visual & 2.8{\scriptsize$\pm$1.0} & 9.2{\scriptsize$\pm$5.5} & \underline{25.0}{\scriptsize$\pm$0.0} & 17.5{\scriptsize$\pm$8.7} & 38.9{\scriptsize$\pm$4.1} & 42.0{\scriptsize$\pm$7.3} & 22.6 \\
    \hspace{1em}AutoMem~\citep{wu2026automemautomatedlearningmemory} & \underline{13.6}{\scriptsize$\pm$4.3} & \underline{41.3}{\scriptsize$\pm$10.5} & 23.8{\scriptsize$\pm$4.0} & \textbf{75.0}{\scriptsize$\pm$19.5} & 40.4{\scriptsize$\pm$7.6} & 31.6{\scriptsize$\pm$4.0} & \underline{37.6} \\
    \rowcolor{oursgray} \hspace{1em}\textbf{EvoViM} & \textbf{100.0}{\scriptsize$\pm$0.0} & \textbf{97.5}{\scriptsize$\pm$4.5} & \textbf{65.0}{\scriptsize$\pm$12.9} & \underline{66.3}{\scriptsize$\pm$30.1} & \underline{45.8}{\scriptsize$\pm$7.5} & 38.6{\scriptsize$\pm$5.9} & \textbf{68.9} \\
    \midrule
    \multicolumn{8}{@{}l}{\textbf{Claude Haiku 5.5}} \\
    \hspace{1em}No memory & 0.3{\scriptsize$\pm$0.6} & 0.0{\scriptsize$\pm$0.0} & 22.5{\scriptsize$\pm$5.3} & \underline{22.5}{\scriptsize$\pm$5.3} & 37.6{\scriptsize$\pm$4.1} & 31.8{\scriptsize$\pm$5.9} & 19.1 \\
    \hspace{1em}Rolling summary & 7.8{\scriptsize$\pm$1.6} & 0.4{\scriptsize$\pm$1.3} & \underline{23.8}{\scriptsize$\pm$4.0} & 18.8{\scriptsize$\pm$10.6} & \underline{41.5}{\scriptsize$\pm$4.1} & 32.9{\scriptsize$\pm$5.6} & 20.9 \\
    \hspace{1em}Hand-crafted text & 7.5{\scriptsize$\pm$1.2} & 22.1{\scriptsize$\pm$10.6} & \underline{23.8}{\scriptsize$\pm$4.0} & 8.8{\scriptsize$\pm$10.3} & 32.9{\scriptsize$\pm$7.3} & \underline{46.3}{\scriptsize$\pm$11.8} & 23.6 \\
    \hspace{1em}Hand-crafted visual & 4.4{\scriptsize$\pm$2.7} & 13.8{\scriptsize$\pm$8.1} & 15.0{\scriptsize$\pm$9.9} & 18.8{\scriptsize$\pm$6.6} & 36.3{\scriptsize$\pm$6.4} & 40.1{\scriptsize$\pm$6.5} & 21.4 \\
    \hspace{1em}AutoMem~\citep{wu2026automemautomatedlearningmemory} & \underline{63.6}{\scriptsize$\pm$12.0} & \underline{42.1}{\scriptsize$\pm$14.5} & \underline{23.8}{\scriptsize$\pm$4.0} & 16.3{\scriptsize$\pm$11.9} & 31.9{\scriptsize$\pm$3.3} & 42.4{\scriptsize$\pm$8.6} & \underline{36.7} \\
    \rowcolor{oursgray} \hspace{1em}\textbf{EvoViM} & \textbf{100.0}{\scriptsize$\pm$0.0} & \textbf{77.1}{\scriptsize$\pm$11.5} & \textbf{98.8}{\scriptsize$\pm$4.0} & \textbf{50.0}{\scriptsize$\pm$25.7} & \textbf{75.0}{\scriptsize$\pm$23.5} & \textbf{61.1}{\scriptsize$\pm$22.8} & \textbf{77.0} \\
    \midrule
    \multicolumn{8}{@{}l}{\textbf{Qwen3.8-27B}} \\
    \hspace{1em}No memory & 0.3{\scriptsize$\pm$0.6} & 0.0{\scriptsize$\pm$0.0} & \underline{25.0}{\scriptsize$\pm$0.0} & \underline{25.0}{\scriptsize$\pm$0.0} & 35.3{\scriptsize$\pm$7.8} & 27.0{\scriptsize$\pm$3.7} & 18.8 \\
    \hspace{1em}Rolling summary & 0.8{\scriptsize$\pm$0.8} & 2.1{\scriptsize$\pm$2.2} & \underline{25.0}{\scriptsize$\pm$0.0} & 22.5{\scriptsize$\pm$5.3} & \underline{46.1}{\scriptsize$\pm$11.6} & 32.7{\scriptsize$\pm$14.1} & 21.5 \\
    \hspace{1em}Hand-crafted text & 0.8{\scriptsize$\pm$1.2} & 16.3{\scriptsize$\pm$9.5} & \underline{25.0}{\scriptsize$\pm$0.0} & 22.5{\scriptsize$\pm$7.9} & 36.6{\scriptsize$\pm$3.3} & 27.4{\scriptsize$\pm$7.8} & 21.4 \\
    \hspace{1em}Hand-crafted visual & 1.1{\scriptsize$\pm$1.2} & 7.9{\scriptsize$\pm$6.9} & 23.8{\scriptsize$\pm$4.0} & 23.8{\scriptsize$\pm$4.0} & 38.3{\scriptsize$\pm$6.7} & 39.1{\scriptsize$\pm$14.7} & 22.3 \\
    \hspace{1em}AutoMem~\citep{wu2026automemautomatedlearningmemory} & \underline{13.3}{\scriptsize$\pm$4.7} & \underline{37.9}{\scriptsize$\pm$6.9} & \underline{25.0}{\scriptsize$\pm$0.0} & 17.5{\scriptsize$\pm$12.1} & 42.0{\scriptsize$\pm$2.5} & \underline{42.1}{\scriptsize$\pm$18.0} & \underline{29.6} \\
    \rowcolor{oursgray} \hspace{1em}\textbf{EvoViM} & \textbf{100.0}{\scriptsize$\pm$0.0} & \textbf{45.4}{\scriptsize$\pm$14.6} & \textbf{96.3}{\scriptsize$\pm$8.4} & \textbf{70.0}{\scriptsize$\pm$22.2} & \textbf{99.1}{\scriptsize$\pm$1.9} & \textbf{52.4}{\scriptsize$\pm$19.8} & \textbf{77.2} \\
    \bottomrule
  \end{tabular}
\end{table*}

\begin{figure}[!t]
  \centering
  \includegraphics[width=\linewidth]{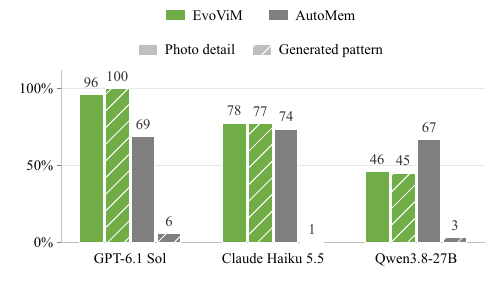}
  \caption{Accuracy on photo details and generated patterns in DMV-Hard.}
  \label{fig:dmv_kind}
\end{figure}

\paragraph{Visual matching.}
On DMV-Hard, EvoViM answers questions about photo details and generated patterns with similar accuracy, while AutoMem succeeds mainly on photo details (Figure~\ref{fig:dmv_kind}).
A matched page directly identifies the product through its URL, so VISTA-style memory answers most questions when a matcher automatically searches its stored frames at every step.

Every EvoViM agent clears Card Matching, where evolved text memory leaves many pairs unmatched.
With Qwen3.8-27B, the flip count on every held-out board matches that of RNG-Bench's greedy policy with perfect memory and row-by-row exploration.

\paragraph{Spatial recall.}
With Claude Haiku 5.5 and Qwen3.8-27B, EvoViM outperforms VISTA-style visual memory on MC Gallery and Poster Maze even when it is given a generic matcher.
The evolved programs link stored views to locations and use these associations to answer queries.
On Poster Maze, the fixed visual memories almost never locate a previously seen poster correctly, whereas EvoViM identifies the correct wall for most of them.
Text memory remains competitive when descriptions are distinctive and stable across observations, as on Poster Maze with GPT-6.1 Sol.
Across the three tasks with deferred queries, EvoViM achieves higher average scores than the fixed visual memories with both models.

\paragraph{Sequential decision-making.}
With Claude Haiku 5.5 and Qwen3.8-27B, EvoViM scores best on both Tetris and Pac-Man.
The evolved programs support decisions from the current frame with computed action suggestions (Figure~\ref{fig:teaser}(c) and (f)).
On Tetris, their placements keep the stack low and earn more points from longer drops.
In Pac-Man, Claude Haiku 5.5 eats on more steps than its baselines, while Qwen3.8-27B spends far fewer steps standing still.
GPT-6.1 Sol does not lead either task, and Section~\ref{sec:exp_failures} examines why its programs provide less support for action selection.

\paragraph{Development and held-out performance.}
Across the 36 final programs of EvoViM and AutoMem, development and held-out scores have a Pearson correlation of 0.89.
Development scores are higher on average, with the largest gaps on Poster Maze and the sequential decision-making tasks.
Among programs that reach perfect development scores, those on Card Matching, MC Gallery, DMV-Hard, and Tetris retain scores of 96--100\% on held-out episodes.
The three such programs on Poster Maze fall to 66--75\%, where different exploration paths can limit the evidence available for recall.

\begin{figure}[!t]
  \centering
  \includegraphics[width=\linewidth]{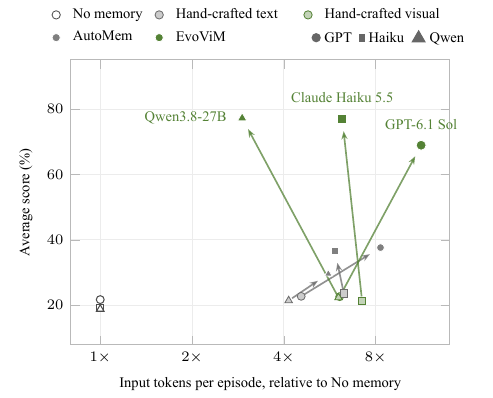}
  \caption{Average score and relative input-token use across the six tasks. Arrows connect each starting program to its evolved version.}
  \label{fig:efficiency}
\end{figure}

\paragraph{Execution cost.}
With Claude Haiku 5.5 and Qwen3.8-27B, the evolved programs achieve higher average scores while using fewer input tokens during execution than the starting program (Figure~\ref{fig:efficiency}).
For this comparison, we normalize token use by No memory on the same model and task, then take the geometric mean across the six tasks.
Compared with AutoMem, EvoViM has similar normalized token use with Claude Haiku 5.5 and roughly half as much with Qwen3.8-27B.
With GPT-6.1 Sol, evolution improves the average score while increasing token use relative to the starting program.

\begin{figure}[!t]
  \centering
  \includegraphics[width=\linewidth]{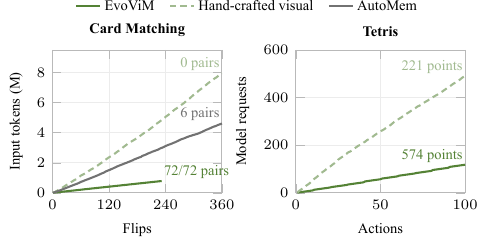}
  \caption{Cumulative input tokens on Card Matching (left) and model requests on Tetris (right), using Qwen3.8-27B on seed 47.}
  \label{fig:tokens}
\end{figure}

The evolved tools perform visual comparisons and action calculations in code, reducing repeated inspection by the model.
Figure~\ref{fig:tokens} shows the execution of Qwen3.8-27B on one held-out seed for Card Matching and Tetris.
In the Card Matching episode, EvoViM clears the board before the flip budget is exhausted and uses about a tenth of the input tokens of its starting program.
Over the same Tetris action budget, it needs roughly a quarter as many model requests.

On DMV-Hard, the every-step matcher searches all stored frames over several scales and rotations, which makes this search costly.
Table~\ref{tab:dmv_cost} reports input tokens and time spent in program code, excluding time waiting for the model.
The evolved programs spend much less time in code than this automatic variant, though they use more model tokens on this task.

\begin{table}[!t]
  \centering
  \footnotesize
  \setlength{\tabcolsep}{3pt}
  \caption{Execution cost on DMV-Hard, averaged over ten episodes. Code time includes matching and excludes waiting for the model.}
  \label{tab:dmv_cost}
  \begin{tabular}{@{}lrrrr@{}}
    \toprule
    & \multicolumn{2}{c}{Claude Haiku 5.5} & \multicolumn{2}{c}{Qwen3.8-27B} \\
    \cmidrule(lr){2-3}\cmidrule(l){4-5}
    Method & Input (M) & Code (s) & Input (M) & Code (s) \\
    \midrule
    VISTA-style archive & 1.53 & 8 & 1.09 & 5 \\
    \quad + matcher (tool) & 1.65 & 1{,}527 & 1.28 & 3{,}585 \\
    \quad + matcher (every step) & 1.49 & 6{,}118 & 1.05 & 5{,}536 \\
    EvoViM & 4.55 & 14 & 2.24 & 364 \\
    \bottomrule
  \end{tabular}
\end{table}

\paragraph{Revision cost.}
Table~\ref{tab:supp_evo} reports the model tokens used during program revision and the number of candidates evaluated, summed over six tasks for each model.
With every model, EvoViM evaluates fewer candidates and accepts more of them.
The recorded input-token totals are similar for the two methods when pooled across models.
For Claude Haiku 5.5 and Qwen3.8-27B, token totals cover only sessions with reported usage, since interrupted sessions do not provide complete counts.

\begin{table}[!t]
  \centering
  \footnotesize
  \setlength{\tabcolsep}{3pt}
  \caption{Candidate evaluations and recorded token use during revision, summed over six tasks for each model. Token counts are in millions.}
  \label{tab:supp_evo}
  \begin{tabular}{@{}lrrrr@{}}
    \toprule
    Method & Evaluated & Accepted & Input & Output \\
    \midrule
    \multicolumn{5}{@{}l}{\textbf{GPT-6.1 Sol}} \\
    AutoMem & 30 & 8 & 85.0 & 1.10 \\
    EvoViM & 26 & 12 & 138.6 & 1.34 \\
    \midrule
    \multicolumn{5}{@{}l}{\textbf{Claude Haiku 5.5}} \\
    AutoMem & 32 & 7 & 508.7 & 9.36 \\
    EvoViM & 21 & 17 & 375.6 & 7.50 \\
    \midrule
    \multicolumn{5}{@{}l}{\textbf{Qwen3.8-27B}} \\
    AutoMem & 30 & 8 & 382.0 & 7.66 \\
    EvoViM & 22 & 17 & 445.1 & 10.34 \\
    \bottomrule
  \end{tabular}
\end{table}

\subsection{Analysis of Evolved Programs}
\label{sec:exp_anatomy}

We compare each final program with its starting program to identify and count the mechanisms added by evolution.
We then examine the development feedback and the model's analysis at every generation to trace how these mechanisms emerged.

\begin{figure}[!t]
  \centering
  \includegraphics[width=\linewidth]{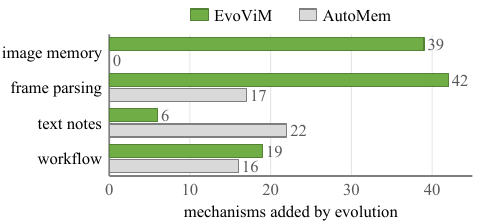}
  \caption{Mechanisms added by evolution across the final programs of six tasks and three models.}
  \label{fig:mechanisms}
\end{figure}

We group the added mechanisms by function, and each program can contribute several mechanisms to a group (Figure~\ref{fig:mechanisms}).
Image memory covers storing, matching, and recalling visual evidence.
The frame-parsing group includes state extraction from visual or textual observations, spatial indexing, and decision support.
Text notes cover records written by the model, while workflow covers changes to prompts or execution flow.

\paragraph{Visual representations and matching.}
Every final EvoViM program reads pixels, with image memory and frame parsing the most frequently added mechanisms.
AutoMem most often adds text notes.
The visual comparisons vary across tasks.
In Card Matching, an exact hash or a pixel difference suffices to identify matching cards, since the two cards of a pair look identical.
Table~\ref{tab:card_programs} compares the card representations and tool interfaces developed by the three models.

\begin{figure}[!t]
  \centering
  \includegraphics[width=\linewidth]{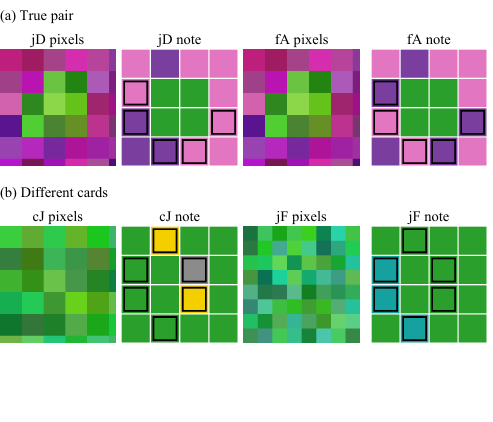}
  \caption{Color-grid notes produced by AutoMem with Claude Haiku 5.5. Boxed cells mark disagreements between the notes. The true pair in (a) agrees in 11 of 16 cells, while the different cards in (b) agree in 10. In both cases, the program ranks the card on the right first.}
  \label{fig:supp_rng_notes}
\end{figure}

The AutoMem programs represent these same cards in text.
GPT-6.1 Sol stores free-text identities, and Qwen3.8-27B stores short color labels.
The Claude Haiku 5.5 program requests a $4{\times}4$ grid of color words for each card and ranks possible partners by agreement between cells (Figure~\ref{fig:supp_rng_notes}).
This grid retains the spatial arrangement of colors, which the shorter descriptions often omit.
Descriptions of matching card faces agree more often than those of different faces, though matching pairs still agree in fewer than half of the cells on average.

\begin{table*}[!t]
  \centering
  \small
  \setlength{\tabcolsep}{5pt}
  \renewcommand{\arraystretch}{1.1}
  \caption{Card representations and tool interfaces in the final programs.}
  \label{tab:card_programs}
  \begin{tabular}{@{}p{0.16\linewidth}p{0.30\linewidth}p{0.47\linewidth}@{}}
    \toprule
    Model & AutoMem & EvoViM \\
    \midrule
    GPT-6.1 Sol
    & Free-text card identities ranked by word overlap.
    & Exact crop hashes. A query tool returns remembered matches when called. \\
    \addlinespace[3pt]
    Claude Haiku 5.5
    & $4{\times}4$ grids of color words ranked by agreement between cells.
    & Crop distances. The program supplies matching-card suggestions and a list of unseen cards. \\
    \addlinespace[3pt]
    Qwen3.8-27B
    & Short color labels compared for equality.
    & Crop distances. The program suggests the next flip at every step. \\
    \bottomrule
  \end{tabular}
\end{table*}

For DMV-Hard and Poster Maze, matching must account for changes in view or lighting.
The DMV-Hard program of GPT-6.1 Sol searches over rotation and scale when comparing a query crop with stored product images.
Its code includes a local-feature matcher resembling SIFT with geometric verification~\citep{lowe2004sift}, although dense correlation supplies most correct matches during execution.
In Poster Maze, Qwen3.8-27B represents each poster as a map of color clusters and compares these maps under the best relabeling and rotation.
Relabeling aligns corresponding regions when the clustering assigns them different color labels.

\begin{figure}[!t]
  \centering
  \includegraphics[width=\linewidth]{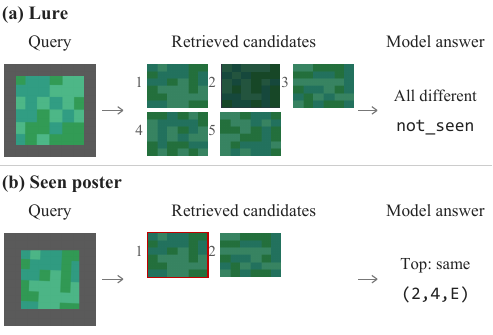}
  \caption{Queries and retrieved candidates from the workspace evolved by GPT-6.1 Sol for Poster Maze, shown during a development episode. Candidates are arranged by rank. The model rejects the lure in (a) and uses the top candidate to locate the seen poster on the east wall of cell (2,4) in (b).}
  \label{fig:case_maze}
\end{figure}

\paragraph{Evidence organization.}
Across the three tasks with deferred queries, the programs index visual evidence by the door, wall, or product requested by later queries.
For each query, the Poster Maze program of GPT-6.1 Sol creates a workspace that keeps the candidate ranking across both execution phases (Figure~\ref{fig:case_maze}).
It shows the query beside the leading stored posters and records the model's verdict for each candidate.
This workspace addresses a failure in an earlier program, where repeated retrieval exhausted the request budget and left lure questions unanswered.
The revised program computes the ranking once and reminds the model to answer on its last request.

\paragraph{Decision support.}
The Card Matching traces show how remembered matches enter the next decision.
GPT-6.1 Sol follows almost every match returned by its query tool, but often skips the query.
Of its 49 missed opportunities to complete a known pair, 48 occur on steps without a query.
Claude Haiku 5.5 follows every explicit matching-card suggestion and chooses almost all remaining flips from the printed list of unseen cards.
The Qwen3.8-27B program suggests the next flip at every step, and the model follows every suggestion throughout its held-out episodes.

\begin{figure*}[!t]
  \centering
  \includegraphics[width=\linewidth]{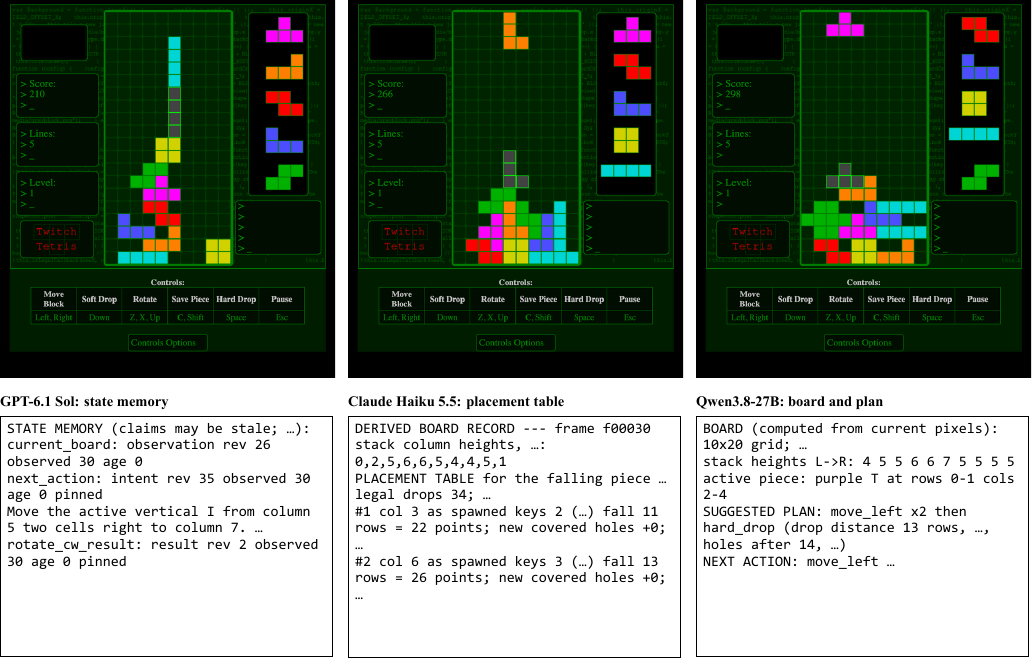}
  \caption{State records and action suggestions produced by the evolved Tetris programs at step 30 of held-out episodes. The excerpts show the information passed to each model.}
  \label{fig:case_tetris}
\end{figure*}

The Tetris programs show similar differences in decision support (Figure~\ref{fig:case_tetris}).
The GPT-6.1 Sol program mainly maintains a ledger of the model's own observations and plans.
Claude Haiku 5.5 receives a board parsed from pixels and a ranking of legal placements.
The Qwen3.8-27B program turns its board estimate into a suggested placement and the next key to press, which the model follows whenever a suggestion is printed.
In Pac-Man, Claude Haiku 5.5 receives computed distances to pellets and ghosts, while the Qwen3.8-27B program recommends a direction that the model follows on almost every step.

Memory can guide the collection of new evidence during exploration.
In Poster Maze, the programs of Claude Haiku 5.5 and Qwen3.8-27B track explored walls and suggest moves toward unexplored areas.
With Qwen3.8-27B, the model follows these suggestions on most navigation steps and sees 4.3 posters clearly per episode, compared with 1.5 for the starting visual program.
The environment can then ask more recall questions about posters encountered during exploration.

\paragraph{Revisions from interaction.}
Across the lineages with deferred queries, the first revision almost always changes what is stored.
The MC Gallery lineage of Claude Haiku 5.5 shows how development failures lead to these changes (Table~\ref{tab:lineage}).
Its starting program keeps notes that never tie a painting to its door.
The first revision adds a pixel matcher and stores painting crops labeled by door and side.
When the model examines the remaining errors, it finds that these crops omit some paintings or capture them only in part.
The next revision therefore stores whole views at each door and determines the side of the painting from its position in the matched view.
This program answers nearly every held-out question.

\begin{table}[!t]
  \centering
  \footnotesize
  \setlength{\tabcolsep}{3pt}
  \renewcommand{\arraystretch}{1.12}
  \caption{The Claude Haiku 5.5 lineage on MC Gallery. Scores are measured on the development episodes after each generation.}
  \label{tab:lineage}
  \begin{tabular}{@{}lcp{0.64\linewidth}@{}}
    \toprule
    Gen. & Score (\%) & Program after selection \\
    \midrule
    0 & 25.0 & The starting program answers \emph{not seen} to every question. \\
    1 & 81.3 & Painting crops indexed by door and side, with a pixel matcher. \\
    2 & 100.0 & Whole views indexed by door, with the side inferred from the match position. \\
    3 & 100.0 & Both candidates are rejected, leaving the generation-2 program unchanged. \\
    \bottomrule
  \end{tabular}
\end{table}

\begin{figure*}[!t]
  \centering
  \includegraphics[width=\linewidth]{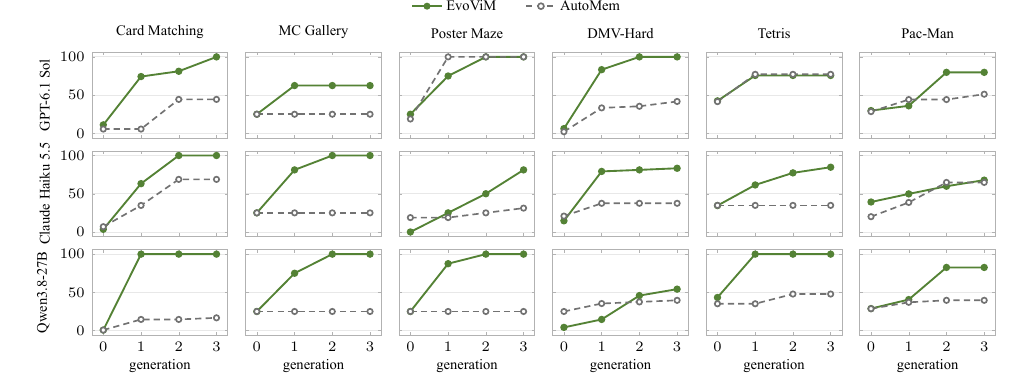}
  \caption{Development scores after each generation, with one row per model. Generation 0 is the starting program.}
  \label{fig:waterfall}
\end{figure*}

Most of the development gain arrives in the first generation (Figure~\ref{fig:waterfall}).
By the second, 13 of the 18 EvoViM lineages have reached at least 90\% of their final gain over the starting program.
Every EvoViM lineage improves at least once, while five AutoMem lineages never accept a candidate.

\paragraph{Reusing revision experience.}
Earlier revisions leave findings and tools that guide later diagnosis and code validation (Table~\ref{tab:revision_experience}).
In the DMV-Hard lineage of Claude Haiku 5.5, the third generation consults the playbook and skips matcher changes that earlier tests found ineffective.
It adds a tool that shows the query beside retrieved candidate regions at the same scale, helping the model compare them directly.
The final program uses this tool throughout its held-out episodes.

\begin{table*}[!t]
  \centering
  \small
  \setlength{\tabcolsep}{5pt}
  \renewcommand{\arraystretch}{1.1}
  \caption{Revision experience reused in the Claude Haiku 5.5 lineage on DMV-Hard.}
  \label{tab:revision_experience}
  \begin{tabular}{@{}p{0.27\linewidth}p{0.31\linewidth}p{0.35\linewidth}@{}}
    \toprule
    Earlier finding & Experience retained & Use in a later revision \\
    \midrule
    A visually plausible candidate was incorrectly used as an answer label.
    & Establish confirmed labels from recorded score changes before testing the matcher.
    & Generations 2 and 3 use score-derived labels to examine matching results. \\
    \addlinespace[3pt]
    Tool declarations need to agree with the interfaces registered by the program.
    & An interface-checking script and a validation procedure in the playbook.
    & Generation 2 adds a script that exports the registered interfaces and runs the inherited checker. \\
    \addlinespace[3pt]
    Several changes to the matcher fail to improve recognition.
    & Notes on the tested changes and guidance to compare aligned candidate regions.
    & Generation 3 consults these records and adds the aligned comparison tool with new rendering checks. \\
    \bottomrule
  \end{tabular}
\end{table*}

By the third generation, the notes contain results from several parent programs.
The model corrects outdated baseline numbers and adds a playbook instruction to read the current parent's episode summary before predicting a candidate's effect.
Across EvoViM lineages, every revision updates its playbook, and nearly half of the later revisions run analysis scripts written during earlier revisions of the same lineage.

\subsection{Failure Analysis}
\label{sec:exp_failures}

\begin{figure}[!t]
  \centering
  \includegraphics[width=\linewidth]{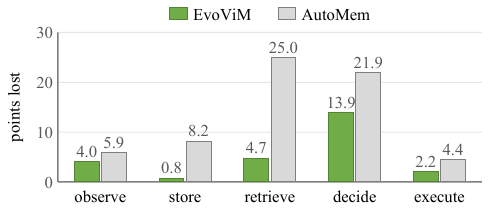}
  \caption{Points lost by AutoMem and EvoViM at each stage, averaged over six tasks and three models.}
  \label{fig:fail_main}
\end{figure}

\paragraph{Attributing score loss.}
To locate where the gains come from, we assign each point lost on a held-out episode to the first stage that failed (Figure~\ref{fig:fail_main}).
We use the recorded frames and memory contents to check observation and storage, asking whether the required evidence was seen clearly and retained.
We then inspect the tool results to determine whether retrieval supplied a usable match.
An incorrect choice based on usable evidence counts as a decision failure, while failure to submit a valid answer or carry out the chosen action falls under execution.
For text memories, a stored entry must identify the relevant location, card cell, or product, with a description where recognition requires one.
If the record exists but its description does not distinguish the item from similar ones, we assign the loss to retrieval.

The allocation rule depends on the task.
In MC Gallery and DMV-Hard, each incorrect answer loses its share of the total score.
Poster Maze also loses points for recall questions that were not asked because the relevant posters were not seen clearly.
For Card Matching, we distribute the score loss from unmatched pairs across stages in proportion to the failed second flips.
For Tetris and Pac-Man, each failed action is weighted by the episode's average points per productive action.
The remaining shortfall is assigned to the decision stage as low-yield play, in which productive actions earn too few points.

\begin{figure*}[!t]
  \centering
  \includegraphics[width=\linewidth]{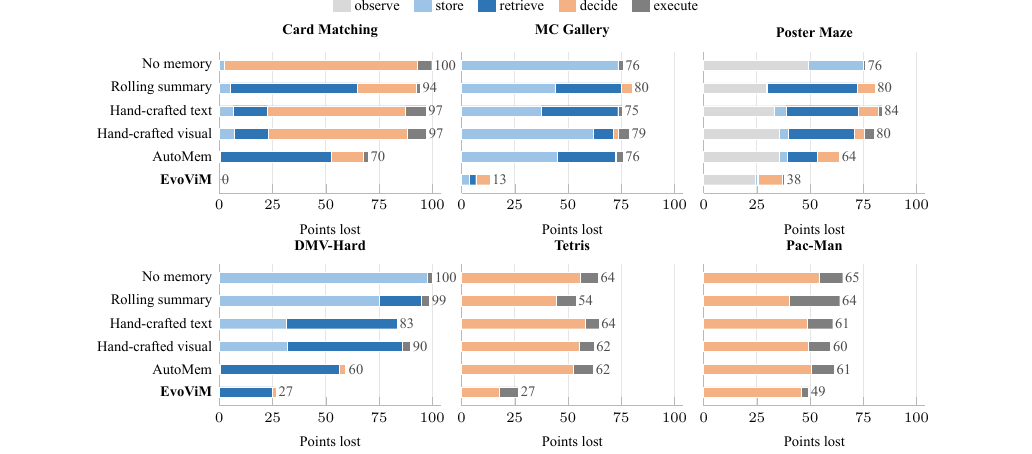}
  \caption{Score loss in each task, averaged over the three models and split by failure stage.}
  \label{fig:fail_games}
\end{figure*}

AutoMem loses most at retrieval, where its descriptions often fail to distinguish visually similar items.
EvoViM's reductions in storage and retrieval failures account for most of its lead over this baseline.
Its remaining losses are concentrated in observation and decision-making (Figure~\ref{fig:fail_games}).

\paragraph{Text retrieval.}
On DMV-Hard, generated patterns share the same palette, and AutoMem's stored descriptions repeatedly use those color words.
In an offline test, we use TF-IDF to rank stored descriptions against each model's description of the query crop.
Across the three models, the median rank of the correct product is 20--28 on pattern questions and two on photo-detail questions.

The color-grid memory of AutoMem with Claude Haiku 5.5 still yields frequent false matches on Card Matching.
When this program labels a candidate as a strong match, fewer than half of the resulting flips complete a pair.
The agent uses the full flip budget in every episode and continues clearing pairs near its end.

\paragraph{Exploration coverage.}
On Poster Maze, the score gap between EvoViM and AutoMem with GPT-6.1 Sol comes from exploration coverage.
Both agents make six answer errors across the ten held-out episodes.
EvoViM receives seven fewer questions, which accounts for the entire score gap between them.
These missing questions occur in two episodes where the agents take different routes at maze junctions.
The EvoViM agent sees only one poster clearly in each of these episodes, leaving little evidence for later recall.

The Qwen3.8-27B program can overestimate its exploration coverage by counting saved side views toward its poster quota.
On four held-out seeds, it marks the quota as met while the environment has registered at most four clearly seen posters.

\begin{figure*}[!t]
  \centering
  \includegraphics[width=\linewidth]{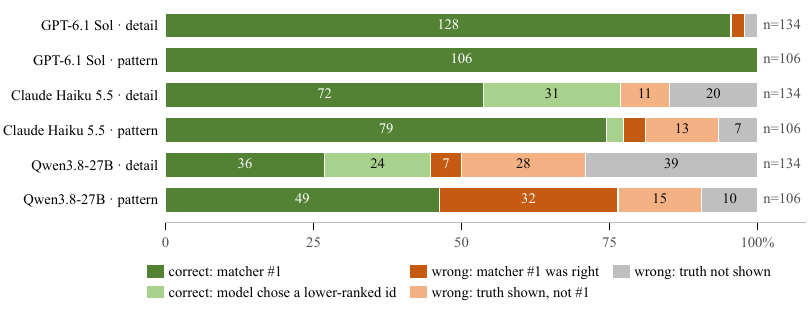}
  \caption{DMV-Hard questions grouped by the rank of the correct product and the final answer. Bar widths show proportions, and segment labels give question counts. For GPT-6.1 Sol, rank 1 includes a first-place result in either the dense or local-feature matcher. For Qwen3.8-27B, it includes a first-place result in any of the three displayed rankings.}
  \label{fig:supp_dmv}
\end{figure*}

\paragraph{Visual retrieval and answer rules.}
Figure~\ref{fig:supp_dmv} compares the rank of the correct product with the final answer on DMV-Hard.
With Qwen3.8-27B, nearly all lost points occur at retrieval, where the correct page ranks below another page or fails the program's confidence threshold.
On some pattern questions, the full-page matcher ranks the correct product first, but the margin rule tells the model not to commit to it.

In the same Qwen3.8-27B program, the candidate pool includes products from the most recent browsing session, even though the questions ask about earlier sessions.
The program also blocks an answer if that product was named in an earlier response.
This treats earlier answers as correct and can exclude the right product after an earlier mistake.

\begin{figure*}[!t]
  \centering
  \includegraphics[width=\linewidth]{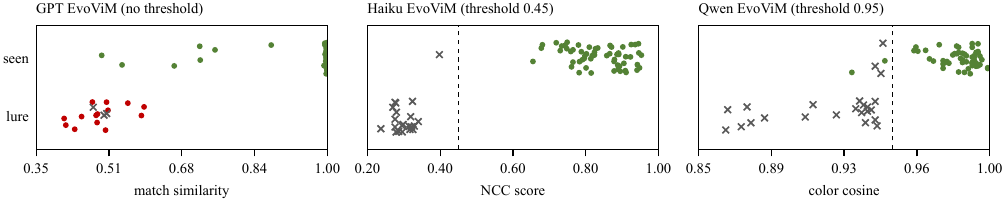}
  \caption{Match scores and answers on MC Gallery. For GPT-6.1 Sol, seen queries are shown only when the true location ranks first. Dots indicate location answers, crosses indicate \emph{not seen}, and dashed lines mark the program's threshold.}
  \label{fig:supp_mc_scores}
\end{figure*}

\paragraph{Decision support.}
Decision failures remain when programs provide evidence without helping the model choose an answer or action.
On MC Gallery, the match scores of GPT-6.1 Sol largely separate seen paintings from lures (Figure~\ref{fig:supp_mc_scores}).
The program has no rule for answering \emph{not seen} when the match score is low.
The model gives a location for 14 of the 17 lures on which it calls the matcher.
Among the final EvoViM programs on MC Gallery and Poster Maze, the three with an explicit rejection rule in code reject every lure.
The GPT-6.1 Sol Poster Maze program leaves this judgment to the model after displaying the candidates and rejects most lures.

\begin{figure}[!t]
  \centering
  \includegraphics[width=\linewidth]{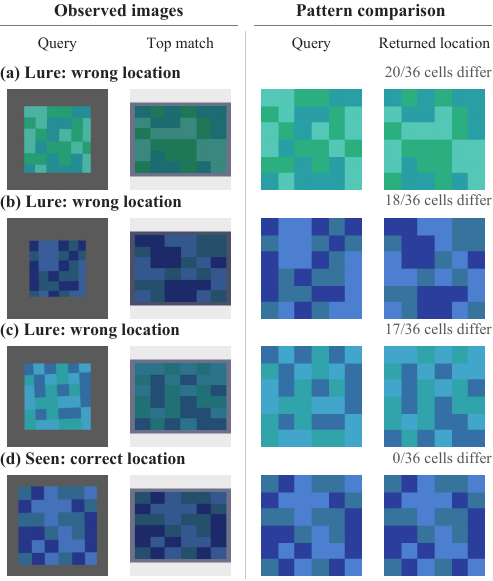}
  \caption{Poster Maze queries answered by Claude Haiku 5.5. The left pair shows the query and the highest-ranked stored view. The right pair shows the underlying patterns for comparison. The model returns incorrect locations for the unseen posters in (a)--(c) and correctly locates the seen poster in (d).}
  \label{fig:supp_maze_lures}
\end{figure}

On Poster Maze, the EvoViM program of Claude Haiku 5.5 returns a wall with every ranked candidate.
Its match scores distinguish lures from seen posters, but the model names a wall for most lures.
These patterns differ visibly from the retrieved posters (Figure~\ref{fig:supp_maze_lures}).

On Tetris and Pac-Man, the accepted GPT-6.1 Sol programs still leave action selection to the model.
In Tetris, its optional grid reader requires the model to specify the board geometry and identify active-piece cells before computing board measurements.
The accepted revisions improve state records but add no code to rank drops or routes.
Their development gains do not translate into comparable improvements on held-out episodes.

\paragraph{Diagnosis during revision.}
In the GPT-6.1 Sol lineage on MC Gallery, the first accepted program explicitly leaves match acceptance and absence decisions to the model.
All four later candidates score below this parent on the development episodes, so the gate retains the earlier program.

In the Claude Haiku 5.5 lineage on Poster Maze, the second revision identifies lure rejection as a gap but cannot determine which failed questions are lures.
A wrong location answer receives zero credit whether the query shows a seen poster or a lure.
No threshold is validated, so the revised program still has no rejection rule.
The Qwen3.8-27B starting program answers every development question with \emph{not seen}, receiving credit only for lures.
These score changes provide labels for checking a rejection threshold.

\subsection{Ablation Study}
\label{sec:ablation}

We ablate the visual pathway and revision support with Claude Haiku 5.5 on MC Gallery and DMV-Hard (Table~\ref{tab:ablation}).
Revision support includes enhanced feedback and the revision experience retained across generations.
To measure variation across runs, we evolve EvoViM and the variant without revision support three times each on MC Gallery, keeping the starting program and development seeds fixed.
Their MC Gallery scores are averaged across these runs, while the other entries report individual lineages.

\begin{table}[!t]
  \centering
  \footnotesize
  \setlength{\tabcolsep}{8pt}
  \caption{Ablation scores (\%) for the visual pathway and revision support with Claude Haiku 5.5.}
  \label{tab:ablation}
  \begin{tabular}{@{}lcc@{}}
    \toprule
    Method & MC Gallery & DMV-Hard \\
    \midrule
    AutoMem & 23.8 & 42.1 \\
    \midrule
    \rowcolor{oursgray} EvoViM & \textbf{88.3} & 77.1 \\
    \quad w/o visual pathway & 22.5 & 42.1 \\
    \quad w/o revision support & 60.0 & \textbf{82.9} \\
    \bottomrule
  \end{tabular}
\end{table}

\paragraph{Visual pathway.}
Removing the visual pathway restricts the program to text memory and removes images from revision.
Enhanced feedback and revision experience remain available, and the agent still sees the current frame during execution.
Scores on both tasks fall to the level of AutoMem.
On MC Gallery, many paintings are never recorded, and descriptions of saved paintings often fit several others.
On DMV-Hard, the text ranker frequently places another product above the correct one.

\paragraph{Feedback and revision experience.}
The variant without revision support diagnoses failures from raw visual trajectories and candidate history.
With revision support, EvoViM improves earlier on both tasks.
On DMV-Hard, its first generation surpasses the variant's final development score after three generations.
By the second generation, all three EvoViM runs on MC Gallery reach at least 75\%, whereas none of the ablated runs reaches this level (Table~\ref{tab:supp_repeat}).
The full setup has a higher mean held-out score on MC Gallery, while the ablated variant finishes slightly higher on DMV-Hard.

\begin{table*}[!t]
  \centering
  \small
  \setlength{\tabcolsep}{8pt}
  \caption{Three evolution runs per configuration on MC Gallery with Claude Haiku 5.5. Development scores are measured after selection at each generation. Held-out columns report accuracy on seen paintings and lures, followed by the overall score. All values are percentages.}
  \label{tab:supp_repeat}
  \begin{tabular}{@{}lrrrrrr@{}}
    \toprule
    & \multicolumn{3}{c}{Development score} & \multicolumn{3}{c}{Held-out performance} \\
    \cmidrule(lr){2-4}\cmidrule(l){5-7}
    Lineage & Gen. 1 & Gen. 2 & Gen. 3 & Seen & Lures & Overall \\
    \midrule
    \multicolumn{7}{@{}l}{\textbf{EvoViM}} \\
    Original & 81.3 & 100.0 & 100.0 & 98.3 & 100.0 & 98.8 \\
    Repeat 1 & 93.8 & 100.0 & 100.0 & 98.3 & 100.0 & 98.8 \\
    Repeat 2 & 31.3 & 75.0 & 75.0 & 90.0 & 0.0 & 67.5 \\
    \midrule
    \multicolumn{7}{@{}l}{\textbf{w/o revision support}} \\
    Original & 50.0 & 50.0 & 50.0 & 5.0 & 50.0 & 16.3 \\
    Repeat 1 & 56.3 & 68.8 & 93.8 & 90.0 & 70.0 & 85.0 \\
    Repeat 2 & 25.0 & 25.0 & 75.0 & 83.3 & 65.0 & 78.8 \\
    \bottomrule
  \end{tabular}
\end{table*}

\paragraph{Variation across lineages.}
All six lineages develop a pixel matcher, but their storage policies and rules for unseen queries differ.
Two full EvoViM lineages answer nearly every held-out question correctly.
The third locates most seen paintings but answers every lure with a location.
A third-generation candidate rejects some development lures but answers fewer seen questions correctly, so the gate retains its parent.
All three full lineages outperform the baselines on MC Gallery.

\begin{figure*}[!t]
  \centering
  \includegraphics[width=\linewidth]{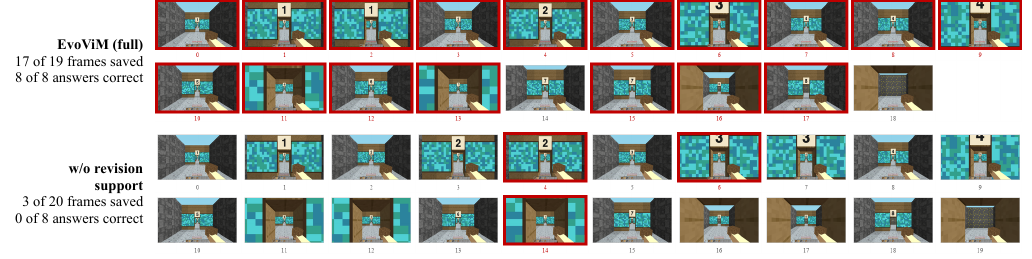}
  \caption{Frames observed on MC Gallery (seed 52) in the original runs with and without revision support. Red outlines mark the frames saved for later retrieval.}
  \label{fig:abl_strip}
\end{figure*}

In the original EvoViM run, the final program requests a view at every door and shows which doors have been covered.
The corresponding run without revision support produces a program that only describes the saving tool, and the model stores few frames (Figure~\ref{fig:abl_strip}).
During revision, the model notices this sparse storage but tests its matcher as though all walk views had been saved.
The two repeated lineages of this variant save views throughout the walk and locate most seen paintings.

\paragraph{Longer evolution for AutoMem.}
We extend the Claude Haiku 5.5 lineage of AutoMem on DMV-Hard from three to five generations.
The fifth generation accepts a candidate that answers one more development question correctly.
Its held-out score is 41.3, close to the 42.1 obtained after three generations.
\section{Conclusion}
\label{sec:conclusion}

In this paper, we studied exploration with deferred queries.
We found that textual descriptions lose visual distinctions, while stored images need task-specific operations to support later decisions.
Programs written from the initial environment instructions alone often prepare for the wrong questions.
To address these problems, we proposed EvoViM, in which a frozen MLLM examines its visual trajectories to identify and revise the operations that lost or misused evidence.
The model also rewrites the playbook and extends the analysis tools used in later revisions, so both the program and the revision process evolve.
We treat memory and skills as two uses of experience carried by this visual program.
Its operations retain evidence for later questions and support action selection from the current state.
Across six tasks and three models, EvoViM achieves the highest average score.
From the same starting program, evolution produces operations suited to each task, with memory and skills combined in programs such as the Card Matching agent that recalls seen cards to suggest the next flip.
We believe EvoViM could extend to embodied environments and tasks that require agents to learn from auditory evidence.

\paragraph{Limitations.}
We have not evaluated EvoViM on tasks involving audio.
Further work is needed to test how it handles sensory noise and combines visual and auditory evidence in embodied environments.
\section{LLM Usage Statement}
Multimodal LLMs are the object of study in this work.
GPT-6.1 Sol, Claude Haiku 5.5, and Qwen3.8-27B perform the tasks and, through the Codex and Claude Code command-line tools, revise the memory programs.
We also used Claude Code as an assistant for the experiment and analysis code, the figures, the search for related work, and the drafting and polishing of the manuscript.
The authors checked all outputs and take full responsibility for the content of this work.
{
    \small
    \bibliographystyle{ieeenat_fullname}
    \bibliography{main}

@String(CVPR= {IEEE Conf. Comput. Vis. Pattern Recog.})

@String(ECCV= {Eur. Conf. Comput. Vis.})

@String(ICLR = {Int. Conf. Learn. Represent.})

@String(CVPR  = {CVPR})

@String(ECCV  = {ECCV})

@String(ICLR  = {ICLR})

@article{Voyager,
  author       = {Guanzhi Wang and
                  Yuqi Xie and
                  Yunfan Jiang and
                  Ajay Mandlekar and
                  Chaowei Xiao and
                  Yuke Zhu and
                  Linxi Fan and
                  Anima Anandkumar},
  title        = {Voyager: An Open-Ended Embodied Agent with Large Language Models},
  journal      = {Trans. Mach. Learn. Res.},
  volume       = {2024},
  year         = {2024},
  url          = {https://openreview.net/forum?id=ehfRiF0R3a},
  bibsource    = {dblp computer science bibliography, https://dblp.org}
}

@inproceedings{Cradle,
  author       = {Weihao Tan and
                  Wentao Zhang and
                  Xinrun Xu and
                  Haochong Xia and
                  Ziluo Ding and
                  Boyu Li and
                  Bohan Zhou and
                  Junpeng Yue and
                  Jiechuan Jiang and
                  Yewen Li and
                  Ruyi An and
                  Molei Qin and
                  Chuqiao Zong and
                  Longtao Zheng and
                  Yujie Wu and
                  Xiaoqiang Chai and
                  Yifei Bi and
                  Tianbao Xie and
                  Pengjie Gu and
                  Xiyun Li and
                  Ceyao Zhang and
                  Long Tian and
                  Chaojie Wang and
                  Xinrun Wang and
                  B{\"{o}}rje F. Karlsson and
                  Bo An and
                  Shuicheng Yan and
                  Zongqing Lu},
  editor       = {Aarti Singh and
                  Maryam Fazel and
                  Daniel Hsu and
                  Simon Lacoste{-}Julien and
                  Felix Berkenkamp and
                  Tegan Maharaj and
                  Kiri Wagstaff and
                  Jerry Zhu},
  title        = {Cradle: Empowering Foundation Agents towards General Computer Control},
  booktitle    = {Forty-second International Conference on Machine Learning, {ICML}
                  2025, Vancouver, BC, Canada, July 13-19, 2025},
  series       = {Proceedings of Machine Learning Research},
  volume       = {267},
  publisher    = {{PMLR} / OpenReview.net},
  year         = {2025},
  url          = {https://proceedings.mlr.press/v267/tan25h.html},
  bibsource    = {dblp computer science bibliography, https://dblp.org}
}

@inproceedings{BALROG,
  author       = {Davide Paglieri and
                  Bartlomiej Cupial and
                  Samuel Coward and
                  Ulyana Piterbarg and
                  Maciej Wolczyk and
                  Akbir Khan and
                  Eduardo Pignatelli and
                  Lukasz Kucinski and
                  Lerrel Pinto and
                  Rob Fergus and
                  Jakob Nicolaus Foerster and
                  Jack Parker{-}Holder and
                  Tim Rockt{\"{a}}schel},
  title        = {{BALROG:} Benchmarking Agentic {LLM} and {VLM} Reasoning On Games},
  booktitle    = {The Thirteenth International Conference on Learning Representations,
                  {ICLR} 2025, Singapore, April 24-28, 2025},
  publisher    = {OpenReview.net},
  year         = {2025},
  url          = {https://openreview.net/forum?id=fp6t3F669F},
  bibsource    = {dblp computer science bibliography, https://dblp.org}
}

@article{VideoGameBench,
  author       = {Alex L. Zhang and
                  Thomas L. Griffiths and
                  Karthik R. Narasimhan and
                  Ofir Press},
  title        = {VideoGameBench: Can Vision-Language Models complete popular video
                  games?},
  journal      = {CoRR},
  volume       = {abs/2505.18134},
  year         = {2025},
  url          = {https://doi.org/10.48550/arXiv.2505.18134},
  doi          = {10.48550/ARXIV.2505.18134},
  eprinttype   = {arXiv},
  eprint       = {2505.18134},
  bibsource    = {dblp computer science bibliography, https://dblp.org}
}

@inproceedings{GameWorld,
  author       = {Mingyu Ouyang and
                  Siyuan Hu and
                  Kevin Qinghong Lin and
                  Hwee Tou Ng and
                  Mike Zheng Shou},
  editor       = {Paolo Favaro and
                  Zuzana Kukelova and
                  Atsuto Maki and
                  Anna Rohrbach and
                  Konrad Schindler and
                  Federico Tombari},
  title        = {GameWorld: Towards Standardized and Verifiable Evaluation of Multimodal
                  Game Agents},
  booktitle    = {Computer Vision - {ECCV} 2026 - 19th European Conference, Malm{\"{o}},
                  Sweden, September 8-12, 2026, Proceedings, Part {LIII}},
  series       = {Lecture Notes in Computer Science},
  volume       = {17053},
  pages        = {41--59},
  publisher    = {Springer},
  year         = {2026},
  url          = {https://doi.org/10.1007/978-3-032-37255-0\_3},
  doi          = {10.1007/978-3-032-37255-0\_3},
  bibsource    = {dblp computer science bibliography, https://dblp.org}
}

@article{JARVIS-1,
  author       = {Zihao Wang and
                  Shaofei Cai and
                  Anji Liu and
                  Yonggang Jin and
                  Jinbing Hou and
                  Bowei Zhang and
                  Haowei Lin and
                  Zhaofeng He and
                  Zilong Zheng and
                  Yaodong Yang and
                  Xiaojian Ma and
                  Yitao Liang},
  title        = {{JARVIS-1:} Open-World Multi-Task Agents With Memory-Augmented Multimodal
                  Language Models},
  journal      = {{IEEE} Trans. Pattern Anal. Mach. Intell.},
  volume       = {47},
  number       = {3},
  pages        = {1894--1907},
  year         = {2025},
  url          = {https://doi.org/10.1109/TPAMI.2024.3511593},
  doi          = {10.1109/TPAMI.2024.3511593},
  bibsource    = {dblp computer science bibliography, https://dblp.org}
}

@inproceedings{MrSteve,
  author       = {Junyeong Park and
                  Junmo Cho and
                  Sungjin Ahn},
  title        = {MrSteve: Instruction-Following Agents in Minecraft with What-Where-When
                  Memory},
  booktitle    = {The Thirteenth International Conference on Learning Representations,
                  {ICLR} 2025, Singapore, April 24-28, 2025},
  publisher    = {OpenReview.net},
  year         = {2025},
  url          = {https://openreview.net/forum?id=CjXaMI2kUH},
  bibsource    = {dblp computer science bibliography, https://dblp.org}
}

@article{AtlasVA,
  author       = {Pan Wang and
                  Yihao Hu and
                  Xiujin Liu and
                  Jingchu Yang and
                  Hang Wang and
                  Zhihao Wen},
  title        = {AtlasVA: Self-Evolving Visual Skill Memory for Teacher-Free {VLM}
                  Agents},
  journal      = {CoRR},
  volume       = {abs/2605.17933},
  year         = {2026},
  url          = {https://doi.org/10.48550/arXiv.2605.17933},
  doi          = {10.48550/ARXIV.2605.17933},
  eprinttype   = {arXiv},
  eprint       = {2605.17933},
  bibsource    = {dblp computer science bibliography, https://dblp.org}
}

@article{rngbench,
  author       = {Shengyuan Ding and
                  Xilin Wei and
                  Xinyu Fang and
                  Haodong Duan and
                  Dahua Lin and
                  Jiaqi Wang and
                  Yuhang Zang},
  title        = {Beyond the Current Observation: Evaluating Multimodal Large Language
                  Models in Controllable Non-Markov Games},
  journal      = {CoRR},
  volume       = {abs/2606.19338},
  year         = {2026},
  url          = {https://doi.org/10.48550/arXiv.2606.19338},
  doi          = {10.48550/ARXIV.2606.19338},
  eprinttype   = {arXiv},
  eprint       = {2606.19338},
  bibsource    = {dblp computer science bibliography, https://dblp.org}
}

@article{memeye,
  author       = {Minghao Guo and
                  Qingyue Jiao and
                  Zeru Shi and
                  Yihao Quan and
                  Boxuan Zhang and
                  Danrui Li and
                  Liwei Che and
                  Wujiang Xu and
                  Shilong Liu and
                  Zirui Liu and
                  Mubbasir Kapadia and
                  Vladimir Pavlovic and
                  Jiang Liu and
                  Mengdi Wang and
                  Yiyu Shi and
                  Dimitris N. Metaxas and
                  Ruixiang Tang},
  title        = {MemEye: {A} Visual-Centric Evaluation Framework for Multimodal Agent
                  Memory},
  journal      = {CoRR},
  volume       = {abs/2605.15128},
  year         = {2026},
  url          = {https://doi.org/10.48550/arXiv.2605.15128},
  doi          = {10.48550/ARXIV.2605.15128},
  eprinttype   = {arXiv},
  eprint       = {2605.15128},
  bibsource    = {dblp computer science bibliography, https://dblp.org}
}

@article{dmvbench,
  author       = {Yujin Tang and
                  Chenming Shang and
                  Ruize Xu and
                  Nikhil Singh},
  title        = {DMV-Bench: Diagnosing Long-Horizon Multimodal Agents' Visual
                  Memory with Incidental Cue Injection},
  journal      = {CoRR},
  volume       = {abs/2606.27499},
  year         = {2026},
  url          = {https://doi.org/10.48550/arXiv.2606.27499},
  doi          = {10.48550/ARXIV.2606.27499},
  eprinttype   = {arXiv},
  eprint       = {2606.27499},
  bibsource    = {dblp computer science bibliography, https://dblp.org}
}

@inproceedings{selfrewarding2024,
  author       = {Weizhe Yuan and
                  Richard Yuanzhe Pang and
                  Kyunghyun Cho and
                  Xian Li and
                  Sainbayar Sukhbaatar and
                  Jing Xu and
                  Jason Weston},
  editor       = {Ruslan Salakhutdinov and
                  Zico Kolter and
                  Katherine A. Heller and
                  Adrian Weller and
                  Nuria Oliver and
                  Jonathan Scarlett and
                  Felix Berkenkamp},
  title        = {Self-Rewarding Language Models},
  booktitle    = {Forty-first International Conference on Machine Learning, {ICML} 2024,
                  Vienna, Austria, July 21-27, 2024},
  series       = {Proceedings of Machine Learning Research},
  volume       = {235},
  pages        = {57905--57923},
  publisher    = {{PMLR} / OpenReview.net},
  year         = {2024},
  url          = {https://proceedings.mlr.press/v235/yuan24d.html},
  bibsource    = {dblp computer science bibliography, https://dblp.org}
}

@inproceedings{absolutezero2025,
  author       = {Andrew Zhao and
                  Yiran Wu and
                  Tong Wu and
                  Quentin Xu and
                  Yang Yue and
                  Matthieu Lin and
                  Shenzhi Wang and
                  Qingyun Wu and
                  Zilong Zheng and
                  Gao Huang},
  editor       = {Danielle Belgrave and
                  Cheng Zhang and
                  Laura N. Montoya and
                  Hsuan{-}Tien Lin and
                  Razvan Pascanu and
                  Piotr Koniusz and
                  Marzyeh Ghassemi and
                  Nancy Chen and
                  Iv{\'{a}}n Vladimir Meza Ru{\'{\i}}z and
                  Arturo Loaiza{-}Bonilla},
  title        = {Absolute Zero: Reinforced Self-play Reasoning with Zero Data},
  booktitle    = {Advances in Neural Information Processing Systems 38: Annual Conference
                  on Neural Information Processing Systems 2025, NeurIPS 2025, San Diego,
                  CA, USA, December 2-7, 2025 / Mexico City, Mexico, November 30 - December
                  5, 2025},
  year         = {2025},
  url          = {http://papers.nips.cc/paper\_files/paper/2025/hash/9837dc00ff67d176373268ed48042d49-Abstract-Conference.html},
  bibsource    = {dblp computer science bibliography, https://dblp.org}
}

@inproceedings{mstar2024,
  author       = {Wei Liu and
                  Junlong Li and
                  Xiwen Zhang and
                  Fan Zhou and
                  Yu Cheng and
                  Junxian He},
  editor       = {Aarti Singh and
                  Maryam Fazel and
                  Daniel Hsu and
                  Simon Lacoste{-}Julien and
                  Felix Berkenkamp and
                  Tegan Maharaj and
                  Kiri Wagstaff and
                  Jerry Zhu},
  title        = {Diving into Self-Evolving Training for Multimodal Reasoning},
  booktitle    = {Forty-second International Conference on Machine Learning, {ICML}
                  2025, Vancouver, BC, Canada, July 13-19, 2025},
  series       = {Proceedings of Machine Learning Research},
  volume       = {267},
  publisher    = {{PMLR} / OpenReview.net},
  year         = {2025},
  url          = {https://proceedings.mlr.press/v267/liu25aj.html},
  bibsource    = {dblp computer science bibliography, https://dblp.org}
}

@inproceedings{
agent0vl2025,
title={Agent0-{VL}: Exploring Self-Evolving Agent for Tool-Integrated Vision-Language Reasoning},
author={Jiaqi Liu and Kaiwen Xiong and Peng Xia and Yiyang Zhou and Haonian Ji and Lu Feng and Siwei Han and Mingyu Ding and Huaxiu Yao},
booktitle={Forty-third International Conference on Machine Learning},
year={2026},
url={https://openreview.net/forum?id=aA02f0y0LY}
}

@article{sima22025,
  author       = {SIMA Team and
                  Google DeepMind},
  title        = {{SIMA} 2: {A} Generalist Embodied Agent for Virtual Worlds},
  journal      = {CoRR},
  volume       = {abs/2512.04797},
  year         = {2025},
  url          = {https://doi.org/10.48550/arXiv.2512.04797},
  doi          = {10.48550/ARXIV.2512.04797},
  eprinttype   = {arXiv},
  eprint       = {2512.04797},
  bibsource    = {dblp computer science bibliography, https://dblp.org}
}

@article{zhou2026hierarchicalselfimprovementframeworktaskspecific,
  author       = {Tailin Zhou},
  title        = {Hierarchical Self-Improvement: {A} Framework for Task-Specific Evolvable
                  Agent Harnesses},
  journal      = {CoRR},
  volume       = {abs/2608.08466},
  year         = {2026},
  url          = {https://doi.org/10.48550/arXiv.2608.08466},
  doi          = {10.48550/ARXIV.2608.08466},
  eprinttype   = {arXiv},
  eprint       = {2608.08466},
  bibsource    = {dblp computer science bibliography, https://dblp.org}
}

@article{xiong2026learningcontinuallylearnmetalearning,
  author       = {Yiming Xiong and
                  Shengran Hu and
                  Jeff Clune},
  title        = {Learning to Continually Learn via Meta-learning Agentic Memory Designs},
  journal      = {CoRR},
  volume       = {abs/2602.07755},
  year         = {2026},
  url          = {https://doi.org/10.48550/arXiv.2602.07755},
  doi          = {10.48550/ARXIV.2602.07755},
  eprinttype   = {arXiv},
  eprint       = {2602.07755},
  bibsource    = {dblp computer science bibliography, https://dblp.org}
}

@article{wu2026automemautomatedlearningmemory,
  author       = {Shengguang Wu and
                  Hao Zhu and
                  Yuhui Zhang and
                  Xiaohan Wang and
                  Serena Yeung{-}Levy},
  title        = {AutoMem: Automated Learning of Memory as a Cognitive Skill},
  journal      = {CoRR},
  volume       = {abs/2607.01224},
  year         = {2026},
  url          = {https://doi.org/10.48550/arXiv.2607.01224},
  doi          = {10.48550/ARXIV.2607.01224},
  eprinttype   = {arXiv},
  eprint       = {2607.01224},
  bibsource    = {dblp computer science bibliography, https://dblp.org}
}

@article{codewithimage2026,
  author       = {Tianze Yang and
                  Liang Wu and
                  Ruitong Sun and
                  Yucheng Shi and
                  Yanqiao Wang and
                  Mayank Darbari and
                  Ninghao Liu and
                  Jin Sun and
                  Liangjie Hong},
  title        = {Self-Evolving Code-with-Image Reasoning},
  journal      = {CoRR},
  volume       = {abs/2608.11292},
  year         = {2026},
  url          = {https://doi.org/10.48550/arXiv.2608.11292},
  doi          = {10.48550/ARXIV.2608.11292},
  eprinttype   = {arXiv},
  eprint       = {2608.11292},
  bibsource    = {dblp computer science bibliography, https://dblp.org}
}

@article{sui2026learninglearnmultimodalexperience,
  author       = {Xingyu Sui and
                  Weixiang Zhao and
                  Yongxin Tang and
                  Yanyan Zhao and
                  Yang Wu and
                  Dandan Tu and
                  Bing Qin},
  title        = {Learning to Learn from Multimodal Experience},
  journal      = {CoRR},
  volume       = {abs/2605.16857},
  year         = {2026},
  url          = {https://doi.org/10.48550/arXiv.2605.16857},
  doi          = {10.48550/ARXIV.2605.16857},
  eprinttype   = {arXiv},
  eprint       = {2605.16857},
  bibsource    = {dblp computer science bibliography, https://dblp.org}
}

@misc{physevo2026,
      title={PhysEvo: Astra Can Act, Let It}, 
      author={Wenqing Tian and Zeyu Zhang and Zhaocheng Liu and Fengwei Liu and Qiang Liu and Liang Wang},
      year={2026},
      eprint={2610.08995},
      archivePrefix={arXiv},
      primaryClass={cs.RO},
      url={https://arxiv.org/abs/2610.08995}, 
}

@article{memevo2026,
  author       = {Guohong Liu and
                  Jialei Ye and
                  Shanhui Zhao and
                  Yunxin Liu and
                  Yuanchun Li},
  title        = {MemEvo: Automatic Discovery of Streaming Video Memory Mechanisms},
  journal      = {CoRR},
  volume       = {abs/2609.36581},
  year         = {2026},
  url          = {https://doi.org/10.48550/arXiv.2609.36581},
  doi          = {10.48550/ARXIV.2609.36581},
  eprinttype   = {arXiv},
  eprint       = {2609.36581},
  bibsource    = {dblp computer science bibliography, https://dblp.org}
}

@misc{han2026vista,
      title={VISTA: A Visual Harness for Reasoning in an Interactive World}, 
      author={Qiushi Han and Keya Hu and Linlu Qiu and Cathy Wu and Kaiming He},
      year={2026},
      eprint={2610.02200},
      archivePrefix={arXiv},
      primaryClass={cs.AI},
      url={https://arxiv.org/abs/2610.02200}, 
}

@article{lowe2004sift,
  author       = {David G. Lowe},
  title        = {Distinctive Image Features from Scale-Invariant Keypoints},
  journal      = {Int. J. Comput. Vis.},
  volume       = {60},
  number       = {2},
  pages        = {91--110},
  year         = {2004},
  url          = {https://doi.org/10.1023/B:VISI.0000029664.99615.94},
  doi          = {10.1023/B:VISI.0000029664.99615.94},
  bibsource    = {dblp computer science bibliography, https://dblp.org}
}

@article{wu2025resum,
  title={Resum: Unlocking long-horizon search intelligence via context summarization},
  author={Wu, Xixi and Li, Kuan and Zhao, Yida and Zhang, Liwen and Ou, Litu and Yin, Huifeng and Zhang, Zhongwang and Yu, Xinmiao and Zhang, Dingchu and Jiang, Yong and others},
  journal={arXiv preprint arXiv:2509.13313},
  year={2025}
}

@misc{openai2026gpt61sol,
  author       = {{OpenAI}},
  title        = {{Addendum to GPT-6 Astra System Card: GPT-6.1 Sol}},
  howpublished = {\url{https://deploymentsafety.openai.com/gpt-6-1-sol}},
  year         = {2026}
}

@misc{anthropic2026haiku55,
  author       = {{Anthropic}},
  title        = {{System Card: Claude Haiku 5.5}},
  howpublished = {\url{https://www.anthropic.com/claude-haiku-5-5-system-card}},
  year         = {2026}
}

@misc{qwen2026qwen38,
  author       = {{Qwen Team}},
  title        = {{Qwen3.8-27B} Model Card},
  howpublished = {\url{https://huggingface.co/Qwen/Qwen3.8-27B}},
  year         = {2026}
}

@misc{anthropic2026claudecode,
  author       = {{Anthropic}},
  title        = {{Claude Code}},
  howpublished = {\url{https://code.claude.com/docs/en/overview}},
  year         = {2026}
}

@misc{openai2026codex,
  author       = {{OpenAI}},
  title        = {{Codex}},
  howpublished = {\url{https://github.com/openai/codex}},
  year         = {2026}
}

@inproceedings{zheng2024seeact,
  author       = {Boyuan Zheng and
                  Boyu Gou and
                  Jihyung Kil and
                  Huan Sun and
                  Yu Su},
  editor       = {Ruslan Salakhutdinov and
                  Zico Kolter and
                  Katherine A. Heller and
                  Adrian Weller and
                  Nuria Oliver and
                  Jonathan Scarlett and
                  Felix Berkenkamp},
  title        = {GPT-4V(ision) is a Generalist Web Agent, if Grounded},
  booktitle    = {Forty-first International Conference on Machine Learning, {ICML} 2024,
                  Vienna, Austria, July 21-27, 2024},
  series       = {Proceedings of Machine Learning Research},
  volume       = {235},
  pages        = {61349--61385},
  publisher    = {{PMLR} / OpenReview.net},
  year         = {2024},
  url          = {https://proceedings.mlr.press/v235/zheng24e.html},
  bibsource    = {dblp computer science bibliography, https://dblp.org}
}

@inproceedings{xie2024osworld,
  author       = {Tianbao Xie and
                  Danyang Zhang and
                  Jixuan Chen and
                  Xiaochuan Li and
                  Siheng Zhao and
                  Ruisheng Cao and
                  Toh Jing Hua and
                  Zhoujun Cheng and
                  Dongchan Shin and
                  Fangyu Lei and
                  Yitao Liu and
                  Yiheng Xu and
                  Shuyan Zhou and
                  Silvio Savarese and
                  Caiming Xiong and
                  Victor Zhong and
                  Tao Yu},
  editor       = {Amir Globersons and
                  Lester Mackey and
                  Danielle Belgrave and
                  Angela Fan and
                  Ulrich Paquet and
                  Jakub M. Tomczak and
                  Cheng Zhang},
  title        = {OSWorld: Benchmarking Multimodal Agents for Open-Ended Tasks in Real
                  Computer Environments},
  booktitle    = {Advances in Neural Information Processing Systems 37: Annual Conference
                  on Neural Information Processing Systems 2024, NeurIPS 2024, Vancouver,
                  BC, Canada, December 10 - 15, 2024},
  year         = {2024},
  url          = {http://papers.nips.cc/paper\_files/paper/2024/hash/5d413e48f84dc61244b6be550f1cd8f5-Abstract-Datasets\_and\_Benchmarks\_Track.html},
  bibsource    = {dblp computer science bibliography, https://dblp.org}
}

@inproceedings{yang20253dmem,
  author       = {Yuncong Yang and
                  Han Yang and
                  Jiachen Zhou and
                  Peihao Chen and
                  Hongxin Zhang and
                  Yilun Du and
                  Chuang Gan},
  title        = {3D-Mem: 3D Scene Memory for Embodied Exploration and Reasoning},
  booktitle    = {{IEEE/CVF} Conference on Computer Vision and Pattern Recognition,
                  {CVPR} 2025, Nashville, TN, USA, June 11-15, 2025},
  pages        = {17294--17303},
  publisher    = {Computer Vision Foundation / {IEEE}},
  year         = {2025},
  url          = {https://openaccess.thecvf.com/content/CVPR2025/html/Yang\_3D-Mem\_3D\_Scene\_Memory\_for\_Embodied\_Exploration\_and\_Reasoning\_CVPR\_2025\_paper.html},
  doi          = {10.1109/CVPR52734.2025.01612},
  bibsource    = {dblp computer science bibliography, https://dblp.org}
}

@inproceedings{park2023generative,
  author       = {Joon Sung Park and
                  Joseph C. O'Brien and
                  Carrie Jun Cai and
                  Meredith Ringel Morris and
                  Percy Liang and
                  Michael S. Bernstein},
  editor       = {Sean Follmer and
                  Jeff Han and
                  J{\"{u}}rgen Steimle and
                  Nathalie Henry Riche},
  title        = {Generative Agents: Interactive Simulacra of Human Behavior},
  booktitle    = {Proceedings of the 36th Annual {ACM} Symposium on User Interface Software
                  and Technology, {UIST} 2023, San Francisco, CA, USA, 29 October 2023-
                  1 November 2023},
  pages        = {2:1--2:22},
  publisher    = {{ACM}},
  year         = {2023},
  url          = {https://doi.org/10.1145/3586183.3606763},
  doi          = {10.1145/3586183.3606763},
  bibsource    = {dblp computer science bibliography, https://dblp.org}
}

@article{packer2023memgpt,
  title={Memgpt: Towards llms as operating systems},
  author={Packer, Charles and Wooders, Sarah and Lin, Kevin and Fang, Vivian and Patil, Shishir G and Stoica, Ion and Gonzalez, Joseph E},
  journal={arXiv preprint arXiv:2310.08560},
  year={2023}
}

@inproceedings{openeqa,
  author       = {Arjun Majumdar and
                  Anurag Ajay and
                  Xiaohan Zhang and
                  Pranav Putta and
                  Sriram Yenamandra and
                  Mikael Henaff and
                  Sneha Silwal and
                  Paul McVay and
                  Oleksandr Maksymets and
                  Sergio Arnaud and
                  Karmesh Yadav and
                  Qiyang Li and
                  Ben Newman and
                  Mohit Sharma and
                  Vincent{-}Pierre Berges and
                  Shiqi Zhang and
                  Pulkit Agrawal and
                  Yonatan Bisk and
                  Dhruv Batra and
                  Mrinal Kalakrishnan and
                  Franziska Meier and
                  Chris Paxton and
                  Alexander Sax and
                  Aravind Rajeswaran},
  title        = {OpenEQA: Embodied Question Answering in the Era of Foundation Models},
  booktitle    = {{IEEE/CVF} Conference on Computer Vision and Pattern Recognition,
                  {CVPR} 2024, Seattle, WA, USA, June 16-22, 2024},
  pages        = {16488--16498},
  publisher    = {{IEEE}},
  year         = {2024},
  url          = {https://doi.org/10.1109/CVPR52733.2024.01560},
  doi          = {10.1109/CVPR52733.2024.01560},
  bibsource    = {dblp computer science bibliography, https://dblp.org}
}

@article{memskill2026,
  author       = {Haozhen Zhang and
                  Quanyu Long and
                  Jianzhu Bao and
                  Tao Feng and
                  Weizhi Zhang and
                  Haodong Yue and
                  Wenya Wang},
  title        = {MemSkill: Learning and Evolving Memory Skills for Self-Evolving Agents},
  journal      = {CoRR},
  volume       = {abs/2602.02474},
  year         = {2026},
  url          = {https://doi.org/10.48550/arXiv.2602.02474},
  doi          = {10.48550/ARXIV.2602.02474},
  eprinttype   = {arXiv},
  eprint       = {2602.02474},
  bibsource    = {dblp computer science bibliography, https://dblp.org}
}

@article{mempro2026,
  author       = {Qingshan Liu and
                  Guoqing Wang and
                  Wen Wu and
                  Jingqi Huang and
                  Xinqi Tao and
                  Dejia Song and
                  Jie Zhou and
                  Liang He},
  title        = {MemPro: Agentic Memory Systems as Evolvable Programs},
  journal      = {CoRR},
  volume       = {abs/2606.00619},
  year         = {2026},
  url          = {https://doi.org/10.48550/arXiv.2606.00619},
  doi          = {10.48550/ARXIV.2606.00619},
  eprinttype   = {arXiv},
  eprint       = {2606.00619},
  bibsource    = {dblp computer science bibliography, https://dblp.org}
}

@inproceedings{
zelikman2024stop,
title={Self-Taught Optimizer ({STOP}): Recursively Self-Improving Code Generation},
author={Eric Zelikman and Eliana Lorch and Lester Mackey and Adam Tauman Kalai},
booktitle={First Conference on Language Modeling},
year={2024},
url={https://openreview.net/forum?id=46Zgqo4QIU}
}

@inproceedings{yin2025godel,
  author={Xunjian Yin and Xinyi Wang and Liangming Pan and Li Lin and Xiaojun Wan and William Yang Wang},
  title={Gödel Agent: A Self-Referential Agent Framework for Recursively Self-Improvement},
  year={2025},
  cdate={1735689600000},
  pages={27890-27913},
  url={https://aclanthology.org/2025.acl-long.1354/},
  booktitle={ACL (1)},
}

@inproceedings{
zhang2026dgm,
title={Darwin G\"odel Machine: Open-Ended Evolution of Self-Improving Agents},
author={Jenny Zhang and Shengran Hu and Cong Lu and Robert Tjarko Lange and Jeff Clune},
booktitle={The Fourteenth International Conference on Learning Representations},
year={2026},
url={https://openreview.net/forum?id=pUpzQZTvGY}
}
}

\clearpage
\appendix

\section{Implementation and Revision Instructions}
\label{sec:supp_impl}
\label{sec:supp_method}

\paragraph{Execution settings.}
The two execution phases are called \texttt{LOG} and \texttt{PLAN} in the implementation, corresponding to \emph{Record} and \emph{Act} in the main text.
The record phase is skipped at the first step, before any action has been taken.
Each phase allows up to 20 model requests per step, with a limit of 4,096 output tokens per request.
Requests time out after 180 seconds, or after 300 seconds for the locally served Qwen3.8-27B.
Qwen3.8-27B uses temperature 0.7, top-$p$ 0.8 and top-$k$ 20.
Candidate code can import the Python standard library, NumPy, SciPy and Pillow, and can access only the public observation fields supplied by the host.

The environment executes the action committed by the model.
This accounts for 27,551 of the 27,555 held-out steps.
In the remaining four steps, all on DMV-Hard, the model committed no action and the host used the default action of the environment.

\paragraph{Revision and code checks.}
Revision runs in Claude Code for Claude Haiku 5.5 and in Codex for GPT-6.1 Sol and Qwen3.8-27B, using the same model as task execution.
Before evaluating a candidate, the host replays recorded observations with a stub model that calls every registered tool.
These checks exercise the tool interfaces and restoration of episode state without making model requests.
A retry starts from the same parent and receives the rejected code together with its failure records.

\paragraph{Selection rules.}
EvoViM accepts a candidate with a higher mean development score.
A tied candidate is accepted only if it succeeds on every development seed and uses fewer steps or tokens without increasing the other.
AutoMem requires a strictly higher mean score and rejects a candidate if the rate of malformed requests or fallback actions rises by more than five percentage points.
Section~\ref{sec:supp_evo} reports how these rules affected candidate selection.

\subsection{Revision Prompt}
\label{sec:supp_prompt}

\begin{figure*}[t]
  \centering
  \includegraphics[width=\linewidth]{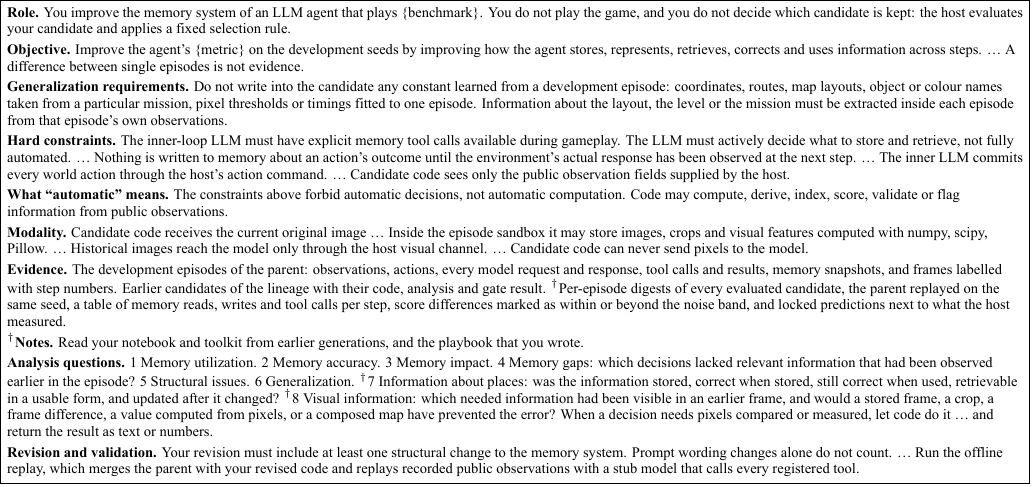}
  \caption{Shared revision-prompt template, abridged. Braces mark fields filled by the host, and \dots{} marks cuts. Parts marked $\dagger$ are removed in the ablation without revision support.}
  \label{fig:supp_prompt}
\end{figure*}

Figure~\ref{fig:supp_prompt} gives the shared revision-prompt template.
It specifies the objective and generalization rules, defines the host interfaces, and provides analysis questions for inspecting the recorded episodes.
The parts marked $\dagger$ form the revision support removed in the corresponding ablation.
They include episode digests and tool-use records, comparisons with earlier predictions, and the revision experience retained in notes, analysis scripts and a playbook.
The additional analysis questions concern spatial associations and the use of visual evidence.
AutoMem receives the description without these marked parts, and its program cannot process pixels or recall stored frames.

The template includes optional parent remeasurement and noise-band reporting, which are disabled in the reported runs.
EvoViM lineage records therefore mark the noise band as not measured.

\subsection{Statistical Analysis}
\label{sec:supp_figs}

We compute 95\% percentile intervals from 10,000 resamples of the held-out seeds.
Comparisons between methods use the same resampled seeds.
For the average across tasks, seeds are resampled within each task and all tasks receive equal weight.
For the mean of three evolution runs, we first average the scores of each seed across runs, then resample the seeds.
These intervals measure variation across held-out seeds.
Variation across evolution runs is reported separately in Table~\ref{tab:supp_repeat}.

\section{Task Details and Additional Measurements}
\label{sec:supp_games}
\label{sec:supp_analysis}

The tables below abbreviate GPT-6.1 Sol, Claude Haiku 5.5 and Qwen3.8-27B as GPT, Haiku and Qwen.
We reconstruct the answer keys from the task seeds and recorded trajectories to analyze individual answers and actions.

\subsection{Card Matching}
\label{sec:supp_rng}

A second flip onto a previously seen card is a recall, and memory precision is the fraction of these recalls that complete a pair.
The card identities reconstructed from the seed agree with all 14,908 logged match outcomes.
Table~\ref{tab:supp_rng} pools recalls across episodes to compute precision and reports the mean number of distinct cells visited per episode.

\begin{table}[t]
  \centering
  \footnotesize
  \setlength{\tabcolsep}{3pt}
  \caption{Card Matching recall precision and exploration coverage. Cells counts distinct cards flipped per episode, out of 144.}
  \label{tab:supp_rng}
  \begin{tabular}{@{}lcccccc@{}}
    \toprule
    & \multicolumn{3}{c}{Precision (\%)} & \multicolumn{3}{c}{Cells} \\
    \cmidrule(lr){2-4}\cmidrule(l){5-7}
    Method & GPT & Haiku & Qwen & GPT & Haiku & Qwen \\
    \midrule
    No memory & 0.0 & 0.1 & 0.0 & 14.8 & 17.0 & 14.8 \\
    Rolling summary & 4.2 & 3.9 & 0.0 & 94.8 & 126.4 & 134.4 \\
    Hand-crafted text & 0.4 & 3.0 & 0.1 & 41.6 & 90.8 & 39.2 \\
    Hand-crafted visual & 1.3 & 1.9 & 0.1 & 46.4 & 83.4 & 24.4 \\
    \midrule
    AutoMem & 7.3 & 34.6 & 8.4 & 132.8 & 144 & 144 \\
    \rowcolor{oursgray} EvoViM & 76.6 & 99.7 & 100 & 144 & 144 & 144 \\
    \bottomrule
  \end{tabular}
\end{table}

\begin{figure*}[t]
  \centering
  \includegraphics[width=\linewidth]{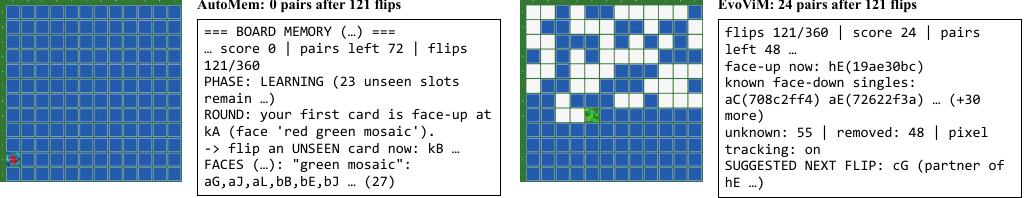}
  \caption{Card Matching at flip 121 of seed 47 with Qwen3.8-27B. The panels show the board and memory presented by AutoMem and EvoViM. Prompt excerpts are quoted verbatim, with \dots{} marking cuts.}
  \label{fig:case_rng}
\end{figure*}

Without evolved exploration rules, some agents repeatedly visit the same few cards.
With Hand-crafted text, GPT-6.1 Sol visits 41.6 distinct cells per episode, while Qwen3.8-27B with Hand-crafted visual visits 24.4.
Rolling summary covers more of the board, but its color descriptions still give a recall precision of at most 4.2\%.
Figure~\ref{fig:case_rng} shows the memory presented to Qwen3.8-27B by the two evolved programs on the same board.
At flip 121, AutoMem has cleared no pair and EvoViM has cleared 24.

The color-grid representation used by AutoMem with Claude Haiku 5.5 continues to find pairs until the flip budget ends.
Its mean number of completed pairs rises from 8.0 after 120 flips to 24.6 after 240 and 45.8 after 360.
The agent takes the highest-ranked partner on 651 of 900 second flips, of which 35\% complete a pair.
When at least 10 of the 16 grid cells agree, the program marks the partner as strong.
The model follows 211 of these 213 suggestions, with 42\% completing a pair.

\subsection{MC Gallery}
\label{sec:supp_mc}

\begin{figure}[t]
  \centering
  \includegraphics[width=\linewidth]{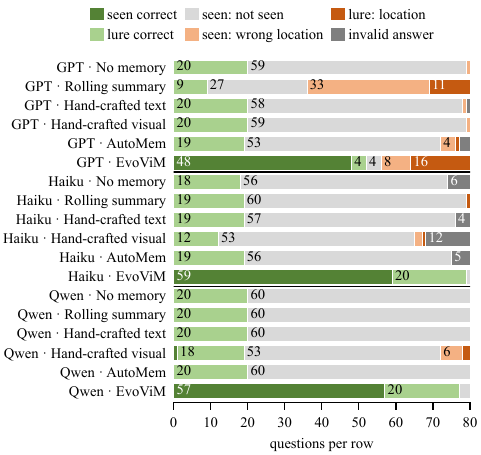}
  \caption{Answers on MC Gallery, separated into seen paintings and lures. Each row contains 60 seen queries and 20 lures over ten episodes.}
  \label{fig:supp_mc_answers}
\end{figure}

Each episode contains six questions about seen paintings and two about lures.
Figure~\ref{fig:supp_mc_answers} separates these outcomes across methods.
Fourteen of the 15 baseline configurations answer \emph{not seen} to 81--100\% of questions.
The exception is GPT-6.1 Sol with Rolling summary, which gives this answer to 45\%.
Across the baseline configurations, only one of 900 seen paintings receives a correct location answer, although every agent passes all 16 paintings in clear view.

\begin{figure*}[t]
  \centering
  \includegraphics[width=\linewidth]{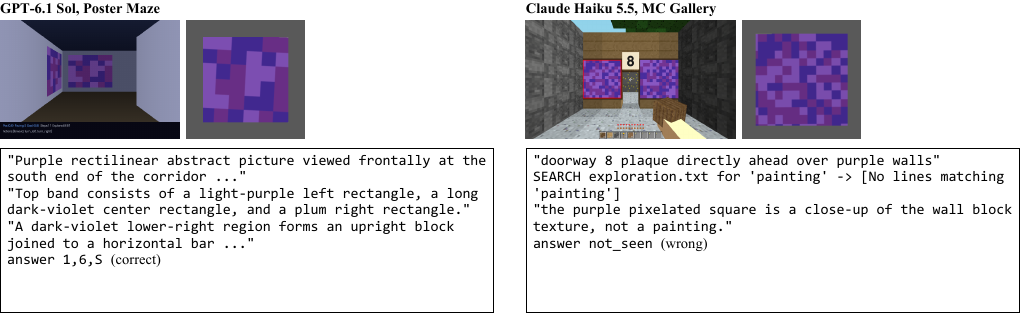}
  \caption{Text memories produced by AutoMem. On Poster Maze, GPT-6.1 Sol records the poster layout and later identifies the correct wall (left). On MC Gallery, Claude Haiku 5.5 records the paintings as walls and later answers \emph{not seen} (right).}
  \label{fig:text_keeps}
\end{figure*}

AutoMem accepts no candidate on MC Gallery with any model, so its final program remains the Hand-crafted text program.
The 18 rejected candidates mostly add text ledgers indexed by doorway and side, but recognize only 9 of 216 seen paintings in development.
One Qwen3.8-27B candidate uses the same description for six different paintings.
Figure~\ref{fig:text_keeps} shows how this loss of distinction differs from a successful text record on Poster Maze.

\subsection{Poster Maze}
\label{sec:supp_maze}

The number of recall questions depends on exploration coverage.
If the agent clearly sees $k$ posters, it receives questions about at most six of them, plus two lures.
The score still uses a denominator of eight, so its upper bound is $100(\min(k,6)+2)/8$.
Table~\ref{tab:supp_maze_nav} reports coverage, while Table~\ref{tab:maze_breakdown} separates answer accuracy from the final score.
Observation counts, question counts and scores are averaged over episodes.
Accuracy is computed over all questions of the corresponding type.

\begin{table}[t]
  \centering
  \footnotesize
  \setlength{\tabcolsep}{3pt}
  \caption{Poster Maze exploration, averaged over ten episodes. Seen counts clearly observed posters. Route is the percentage of cells on the shortest route that the agent visits.}
  \label{tab:supp_maze_nav}
  \begin{tabular}{@{}lcccccc@{}}
    \toprule
    & \multicolumn{2}{c}{GPT} & \multicolumn{2}{c}{Haiku} & \multicolumn{2}{c}{Qwen} \\
    \cmidrule(lr){2-3}\cmidrule(lr){4-5}\cmidrule(l){6-7}
    Method & Seen & Route & Seen & Route & Seen & Route \\
    \midrule
    No memory & 2.1 & 49 & 2.0 & 47 & 2.1 & 41 \\
    Rolling summary & 4.9 & 79 & 4.1 & 65 & 2.6 & 37 \\
    Hand-crafted text & 5.6 & 81 & 3.7 & 49 & 1.9 & 30 \\
    Hand-crafted visual & 4.9 & 79 & 4.0 & 59 & 1.5 & 37 \\
    \midrule
    AutoMem & 5.3 & 81 & 3.5 & 76 & 1.8 & 34 \\
    \rowcolor{oursgray} EvoViM & 4.6 & 76 & 5.5 & 73 & 4.3 & 70 \\
    \bottomrule
  \end{tabular}
\end{table}

\begin{table}[t]
  \centering
  \footnotesize
  \setlength{\tabcolsep}{2pt}
  \caption{Poster Maze questions and answers. Observed and Asked are mean counts per episode. Accuracy columns and Score are percentages.}
  \label{tab:maze_breakdown}
  \begin{tabular}{@{}lcccccc@{}}
    \toprule
    Method & Observed & Asked & All & Seen & Lure & Score \\
    \midrule
    \multicolumn{7}{@{}l}{\textbf{GPT-6.1 Sol}} \\
    \hspace{0.5em}AutoMem & 5.3 & 6.6 & 90.9 & 93.5 & 85.0 & 75.0 \\
    \rowcolor{oursgray} \hspace{0.5em}EvoViM & 4.6 & 5.9 & 89.8 & 97.4 & 75.0 & 66.3 \\
    \multicolumn{7}{@{}l}{\textbf{Claude Haiku 5.5}} \\
    \hspace{0.5em}AutoMem & 3.5 & 5.1 & 25.5 & 9.7 & 50.0 & 16.3 \\
    \rowcolor{oursgray} \hspace{0.5em}EvoViM & 5.5 & 6.4 & 62.5 & 84.1 & 15.0 & 50.0 \\
    \multicolumn{7}{@{}l}{\textbf{Qwen3.8-27B}} \\
    \hspace{0.5em}AutoMem & 1.8 & 3.8 & 36.8 & 0.0 & 70.0 & 17.5 \\
    \rowcolor{oursgray} \hspace{0.5em}EvoViM & 4.3 & 5.9 & 94.9 & 92.3 & 100.0 & 70.0 \\
    \bottomrule
  \end{tabular}
\end{table}

\begin{figure}[t]
  \centering
  \includegraphics[width=\linewidth]{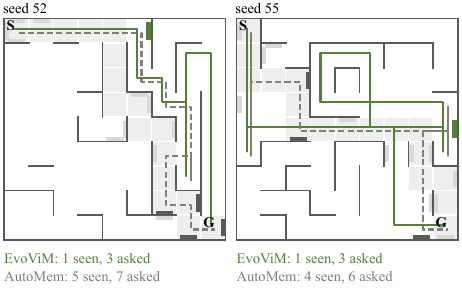}
  \caption{GPT-6.1 Sol routes on Poster Maze seeds 52 and 55. Green solid lines show EvoViM and gray dashed lines show AutoMem. Shading marks the shortest route from S to G. Bars mark posters, with dark bars seen clearly by AutoMem and green outlines seen clearly by EvoViM.}
  \label{fig:supp_maze_routes}
\end{figure}

For GPT-6.1 Sol, the difference in the number of questions comes from seeds 52 and 55.
EvoViM receives three questions on each, while AutoMem receives seven and six.
Figure~\ref{fig:supp_maze_routes} shows where their routes diverge.
The paths have similar lengths, but EvoViM sees only one poster clearly in each maze.
On seed 44, no agent sees more than two posters clearly, which limits every method to a score of 50.

\subsection{DMV-Hard}
\label{sec:supp_dmv}

The ten evaluation episodes contain 134 questions about photo details and 106 about generated patterns.
We reconstruct each query from the task seed and check its answer against the change in the running score.
Table~\ref{tab:supp_dmv} pools the correct answers across episodes for each kind of query.
Figure~\ref{fig:dmv_kind} instead averages the accuracy of each kind over seeds, giving each episode equal weight.

\begin{table}[t]
  \centering
  \footnotesize
  \setlength{\tabcolsep}{3pt}
  \caption{Correct DMV-Hard answers over ten episodes. D denotes the 134 photo-detail questions and P the 106 pattern questions.}
  \label{tab:supp_dmv}
  \begin{tabular}{@{}lcccccc@{}}
    \toprule
    & \multicolumn{2}{c}{GPT} & \multicolumn{2}{c}{Haiku} & \multicolumn{2}{c}{Qwen} \\
    \cmidrule(lr){2-3}\cmidrule(lr){4-5}\cmidrule(l){6-7}
    Method & D & P & D & P & D & P \\
    \midrule
    No memory & 0 & 0 & 0 & 0 & 0 & 0 \\
    Rolling summary & 2 & 2 & 1 & 0 & 2 & 3 \\
    Hand-crafted text & 24 & 3 & 51 & 2 & 38 & 1 \\
    Hand-crafted visual & 20 & 2 & 33 & 0 & 17 & 2 \\
    \midrule
    AutoMem & 92 & 7 & 100 & 1 & 87 & 4 \\
    \rowcolor{oursgray} EvoViM & 128 & 106 & 103 & 82 & 60 & 49 \\
    \bottomrule
  \end{tabular}
\end{table}

Keeping a product image does not always preserve its identity.
Hand-crafted visual stores frames under text labels but does not automatically associate each frame with the page ID.
Across the three models, it recalls 581, 329 and 679 frames, yet views the correct page in only 11, 7 and 24 of the 240 questions.
All but two of its 74 correct answers already have the true product ID in text memory.

\subsection{Tetris and Pac-Man}
\label{sec:supp_arcade}

\paragraph{Tetris scoring.}
GameWorld awards two points per row of a hard drop, one point per soft-drop update, and 100, 300, 500 or 800 points times the level for clearing one to four lines at once.
Task progress is the score of the best game in the episode divided by 600 and capped at one.
Table~\ref{tab:tetris} separates dropping pieces from clearing lines.
Line clears account for 9\% and 12\% of the points earned by EvoViM with Claude Haiku 5.5 and Qwen3.8-27B.
Most points therefore come from keeping the stack low enough for long drops.

\begin{table}[t]
  \centering
  \footnotesize
  \setlength{\tabcolsep}{2.5pt}
  \caption{Tetris statistics over five episodes of 100 actions. Lines and Pieces refer to the best game. Drops is the percentage of its points earned by dropping pieces, and Top-outs counts games lost per episode.}
  \label{tab:tetris}
  \begin{tabular}{@{}lccccc@{}}
    \toprule
    Method & Score & Lines & Drops (\%) & Pieces & Top-outs \\
    \midrule
    \multicolumn{6}{@{}l}{\textbf{GPT-6.1 Sol}} \\
    \hspace{0.5em}No memory & 34.7 & 0.0 & 100 & 12.6 & 1.2 \\
    \hspace{0.5em}Hand-crafted text & 36.9 & 0.0 & 100 & 12.6 & 1.8 \\
    \hspace{0.5em}Hand-crafted visual & 38.9 & 0.0 & 100 & 14.4 & 1.8 \\
    \midrule
    \hspace{0.5em}AutoMem & 40.4 & 0.0 & 100 & 15.6 & 1.4 \\
    \rowcolor{oursgray} \hspace{0.5em}EvoViM & 45.8 & 0.0 & 100 & 16.8 & 1.0 \\
    \multicolumn{6}{@{}l}{\textbf{Claude Haiku 5.5}} \\
    \hspace{0.5em}No memory & 37.6 & 0.0 & 100 & 13.8 & 1.4 \\
    \hspace{0.5em}Hand-crafted text & 32.9 & 0.0 & 100 & 16.8 & 0.8 \\
    \hspace{0.5em}Hand-crafted visual & 36.3 & 0.0 & 100 & 17.0 & 0.8 \\
    \midrule
    \hspace{0.5em}AutoMem & 31.9 & 0.0 & 100 & 13.2 & 0.6 \\
    \rowcolor{oursgray} \hspace{0.5em}EvoViM & 75.0 & 1.0 & 91 & 23.8 & 0.2 \\
    \multicolumn{6}{@{}l}{\textbf{Qwen3.8-27B}} \\
    \hspace{0.5em}No memory & 35.3 & 0.0 & 100 & 17.2 & 0.8 \\
    \hspace{0.5em}Hand-crafted text & 36.6 & 0.0 & 100 & 12.6 & 1.6 \\
    \hspace{0.5em}Hand-crafted visual & 38.3 & 0.0 & 100 & 12.4 & 1.4 \\
    \midrule
    \hspace{0.5em}AutoMem & 42.0 & 0.0 & 100 & 14.0 & 2.4 \\
    \rowcolor{oursgray} \hspace{0.5em}EvoViM & 99.1 & 0.8 & 88 & 23.0 & 0.0 \\
    \bottomrule
  \end{tabular}
\end{table}

\begin{figure*}[t]
  \centering
  \includegraphics[width=\linewidth]{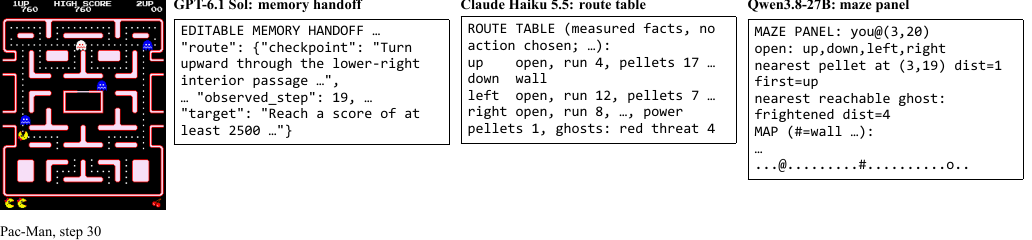}
  \caption{Information presented by the evolved Pac-Man programs. GPT-6.1 Sol retains route claims written by the model. Claude Haiku 5.5 and Qwen3.8-27B compute nearby pellets and ghost distances from the current frame. Prompt excerpts are quoted verbatim, with \dots{} marking cuts.}
  \label{fig:case_arcade}
\end{figure*}

\paragraph{Pac-Man timing and scoring.}
The first maze contains 2,400 points in dots and power pellets, while the target score is 2,500.
Reaching the target therefore requires points from ghosts, fruit or a later level.
No evaluated episode clears the first maze, and only one reaches the target.
Table~\ref{tab:supp_pacman} reports productive and stationary steps.
With Claude Haiku 5.5 and Qwen3.8-27B, EvoViM increases the number of steps on which Pac-Man eats something.
Figure~\ref{fig:case_arcade} shows the information their programs provide for these decisions.

\begin{table}[t]
  \centering
  \footnotesize
  \setlength{\tabcolsep}{3pt}
  \caption{Pac-Man actions, averaged over five episodes of 100 steps. Eat counts steps on which Pac-Man eats something, and Still counts steps without movement.}
  \label{tab:supp_pacman}
  \begin{tabular}{@{}lcccccc@{}}
    \toprule
    & \multicolumn{2}{c}{GPT} & \multicolumn{2}{c}{Haiku} & \multicolumn{2}{c}{Qwen} \\
    \cmidrule(lr){2-3}\cmidrule(lr){4-5}\cmidrule(l){6-7}
    Method & Eat & Still & Eat & Still & Eat & Still \\
    \midrule
    No memory & 34.4 & 3.8 & 29.4 & 8.2 & 22.8 & 18.6 \\
    Rolling summary & 31.6 & 6.6 & 25.4 & 25.0 & 27.4 & 36.2 \\
    Hand-crafted text & 36.0 & 5.0 & 37.4 & 3.2 & 24.4 & 24.8 \\
    Hand-crafted visual & 33.2 & 5.2 & 33.2 & 7.2 & 31.2 & 16.4 \\
    \midrule
    AutoMem & 24.6 & 2.6 & 35.8 & 8.4 & 30.0 & 20.0 \\
    \rowcolor{oursgray} EvoViM & 32.8 & 4.6 & 46.0 & 3.0 & 38.2 & 1.8 \\
    \bottomrule
  \end{tabular}
\end{table}

Pac-Man uses the seed only for the turns of frightened ghosts.
Its first frame is pixel-identical in 89 of the 90 evaluated episodes.
The game advances with wall-clock time, so identical action sequences usually diverge within four to six frames, even with the same seed.
Differences between runs thus include both model sampling and execution timing.

\section{Program Mechanisms}
\label{sec:supp_programs}

\begin{figure}[t]
  \centering
  \includegraphics[width=\linewidth]{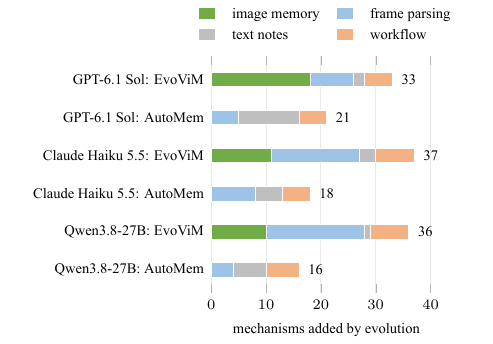}
  \caption{Mechanisms added by each model, grouped as in Figure~\ref{fig:mechanisms} and counted over the six final programs of each method.}
  \label{fig:tools_backbone}
\end{figure}

We compare each of the 36 final programs with its starting program and assign the added mechanisms to the eight functions in Table~\ref{tab:functions}.
The table counts programs that gain each function.
Figures~\ref{fig:mechanisms} and~\ref{fig:tools_backbone} instead count individual mechanisms, so a program can contribute more than once to a group.

\begin{table*}[t]
  \centering
  \footnotesize
  \setlength{\tabcolsep}{4pt}
  \caption{Functions added by evolution. Counts give the number of final programs with each function, out of 18 per method. Group names match Figure~\ref{fig:mechanisms}.}
  \label{tab:functions}
  \begin{tabular}{@{}llp{0.52\linewidth}cc@{}}
    \toprule
    Function & Group & What the added code does & EvoViM & AutoMem \\
    \midrule
    Pixel matching & image memory & compares a query image or region with stored pixels or pixel features and returns a score or a ranking & 14 & 0 \\
    Visual store & image memory & keeps frames, crops or pixel features across steps, filed under a place, an object or an id & 12 & 0 \\
    Visual recall & image memory & lets a tool ask the host to show stored frames or crops to the model & 11 & 0 \\
    Frame parsing & frame parsing & reads the current frame or observation into structured values such as card cells, posters or a board & 13 & 1 \\
    Spatial index & frame parsing & keeps where things are and how places connect, such as a maze graph or place records & 8 & 6 \\
    Decision support & frame parsing & turns stored or parsed evidence into a proposal, such as seen or not seen, the next flip or a ranked drop & 12 & 7 \\
    Text ledger & text notes & keeps records written by the model, with revisions or pinned entries & 6 & 13 \\
    Workflow control & workflow & controls requests within the phase budget, adds reminders, or changes when the model is called & 16 & 13 \\
    \bottomrule
  \end{tabular}
\end{table*}

\begin{figure*}[t]
  \centering
  \includegraphics[width=\linewidth]{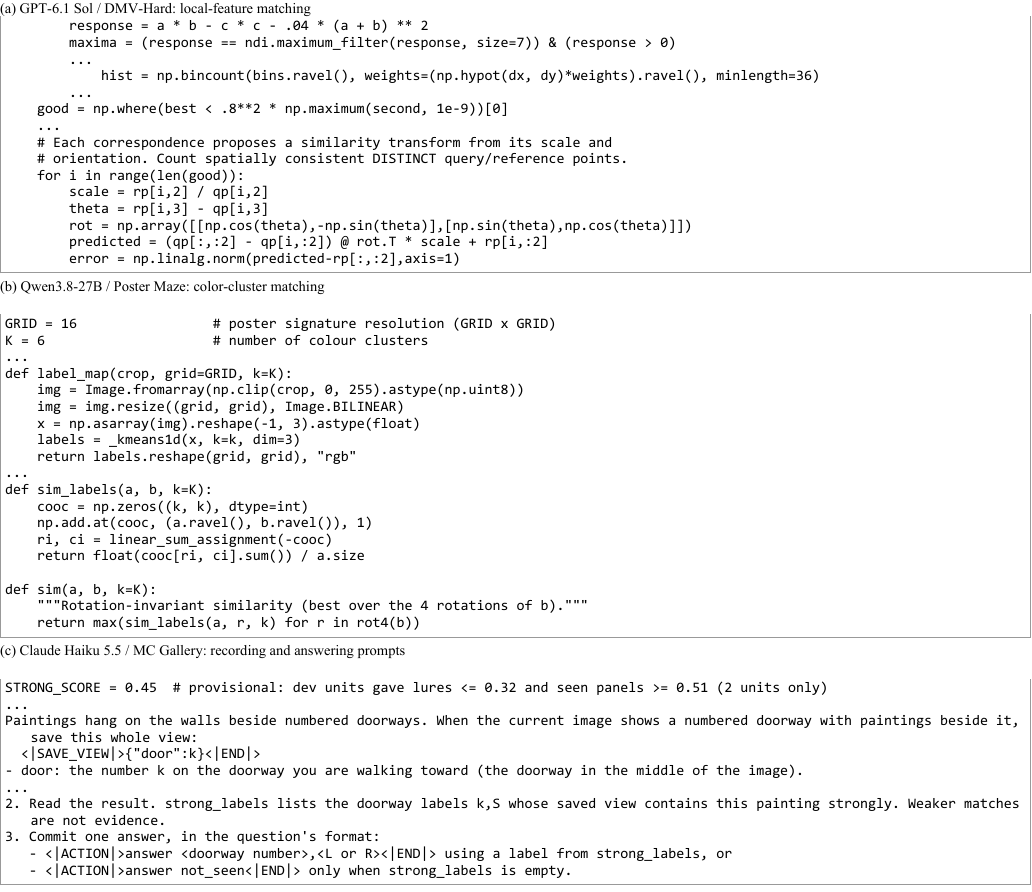}
  \caption{Excerpts of three final EvoViM programs, quoted verbatim. \dots{} marks cuts.}
  \label{fig:supp_code}
\end{figure*}

Figure~\ref{fig:supp_code} gives code excerpts from three final programs.
The DMV-Hard program of GPT-6.1 Sol detects Harris corners at several scales and describes them with gradient-orientation histograms.
It filters matches by a ratio test and checks whether they agree on a similarity transform.
The Poster Maze program of Qwen3.8-27B compares $16\times16$ maps of six color clusters under relabeling and four rotations.
The MC Gallery program of Claude Haiku 5.5 associates stored views with doorways and requires a strong visual match before returning a location.

\section{Failure Attribution}
\label{sec:supp_failures}

Table~\ref{tab:fail_taxonomy} defines the subtypes used within the five failure stages of Section~\ref{sec:exp_failures}.
We inspect the memory files, prompts and decision frames, with at least two verbatim spot checks per subtype.
Table~\ref{tab:fail_sub} reports the resulting loss breakdown for AutoMem and EvoViM.

For MC Gallery, Poster Maze and DMV-Hard, each incorrect answer contributes its share of the task score.
For Card Matching, the second flip is treated as a query about the mate of the first card.
An episode's score loss is divided in proportion to its failed flips.
Tetris and Pac-Man have no recall stage, so their losses fall under decision or execution.
Each failed action is assigned the points earned per productive action in that episode, with the remaining shortfall assigned to low-yield play.

For text memory, an item counts as stored when an entry identifies its doorway, wall, cell or product ID.
MC Gallery and Card Matching additionally require a description of the item.
This criterion determines the split between storage and retrieval losses, while their sum remains unchanged.
Of the 12,118 failed Card Matching flips, 64 have unresolved storage subtypes because their text notes use cell notation that cannot be mapped to the board.

\begin{table*}[t]
  \centering
  \footnotesize
  \setlength{\tabcolsep}{4pt}
  \caption{Failure subtypes and their assignment criteria. Tasks lists where each subtype occurs in the evaluated methods.}
  \label{tab:fail_taxonomy}
  \begin{tabular}{@{}llp{0.43\linewidth}l@{}}
    \toprule
    Stage & Subtype & Criterion & Tasks \\
    \midrule
    Observe & never on screen & the item did not appear on screen & Maze \\
     & no clear view & it appeared, but not clearly or head-on & Maze \\
    \midrule
    Store & not saved & no memory entry holds the item at the decision step & Card, MC, Maze, DMV \\
     & saved under a wrong key & an entry exists but is bound to a wrong doorway, wall, cell or id & Card, MC, Maze, DMV \\
     & saved but incomplete & the entry lacks what the answer needs: the id of a saved photo, the painting in a labeled view, or a usable face description & Card, MC, DMV \\
    \midrule
    Retrieve & not queried & a usable entry exists, but no retrieval was made or shown at the decision step & Card, MC, Maze, DMV \\
     & ranked below another item & a ranking was shown, and the true item was not first & Card, MC, Maze, DMV \\
     & ranked first, below threshold & the true item was first, but a rule of the program declared no match & Card, MC, Maze, DMV \\
     & words fit other items & a text entry for the item was shown, but its words also fit other items & Card, MC, Maze, DMV \\
    \midrule
    Decide & match ignored & the top result shown was the true item, and the model answered \emph{not seen} or explored & Card, MC, Maze, DMV \\
     & mapped to a wrong answer & the top result shown was the true item, and the model answered a wrong location, side or id & Card, MC, Maze, DMV \\
     & lure: program named a place & a lure was answered with the location that the program offered & Card, MC, Maze \\
     & lure: model claimed a place & a lure was answered as seen without such an offer, or, in Card Matching, a flip onto a known card whose mate was not seen & Card, MC, Maze \\
     & stack topped out & actions in Tetris games that topped out and do not set the score & Tetris \\
     & caught by a ghost & Pac-Man actions from a collision to the first move after respawn & Pac-Man \\
     & unproductive moves & Tetris moves that cancel out, waits and extra rotations, and Pac-Man moves that eat nothing although a dot was within reach & Tetris, Pac-Man \\
     & low-yield play & points still missing if every action had earned the points per productive action of that episode & Tetris, Pac-Man \\
    \midrule
    Execute & invalid answer & a malformed answer or a non-answer action at a question & MC, Maze, DMV \\
     & invalid action & an illegal flip onto a removed card or the same card & Card \\
     & idle action & a valid action without effect, such as a move into a wall, an ignored drop or standing still & Tetris, Pac-Man \\
     & no action & the request cap was reached and the host fell back & Maze, DMV, Tetris, Pac-Man \\
    \bottomrule
  \end{tabular}
\end{table*}

\begin{table*}[t]
  \centering
  \scriptsize
  \setlength{\tabcolsep}{2.6pt}
  \caption{Points lost by failure subtype, averaged over the three models. A denotes AutoMem and E denotes EvoViM. All averages over the six tasks. A dash indicates fewer than 0.05 points.}
  \label{tab:fail_sub}
  \begin{tabular}{@{}llrrrrrrrrrrrrrr@{}}
    \toprule
     & & \multicolumn{2}{c}{Card} & \multicolumn{2}{c}{MC} & \multicolumn{2}{c}{Maze} & \multicolumn{2}{c}{DMV} & \multicolumn{2}{c}{Tetris} & \multicolumn{2}{c}{Pac-Man} & \multicolumn{2}{c}{All} \\
    \cmidrule(lr){3-4}\cmidrule(lr){5-6}\cmidrule(lr){7-8}\cmidrule(lr){9-10}\cmidrule(lr){11-12}\cmidrule(lr){13-14}\cmidrule(lr){15-16}
    Stage & Subtype & A & E & A & E & A & E & A & E & A & E & A & E & A & E \\
    \midrule
    Observe & never on screen & -- & -- & -- & -- & 13.3 & 5.0 & -- & -- & -- & -- & -- & -- & 2.2 & 0.8 \\
     & no clear view & -- & -- & -- & -- & 22.1 & 19.2 & -- & -- & -- & -- & -- & -- & 3.7 & 3.2 \\
    \addlinespace[2pt]
    Store & not saved & -- & -- & 45.0 & -- & 3.8 & -- & 0.1 & -- & -- & -- & -- & -- & 8.1 & -- \\
     & saved under a wrong key & -- & -- & -- & -- & -- & 0.8 & -- & -- & -- & -- & -- & -- & -- & 0.1 \\
     & saved but incomplete & 0.2 & -- & -- & 3.8 & -- & -- & -- & -- & -- & -- & -- & -- & -- & 0.6 \\
    \addlinespace[2pt]
    Retrieve & not queried & -- & -- & -- & 1.7 & -- & -- & 0.6 & -- & -- & -- & -- & -- & 0.1 & 0.3 \\
     & ranked below another item & 51.5 & -- & -- & -- & 10.8 & 0.4 & 55.7 & 20.0 & -- & -- & -- & -- & 19.7 & 3.4 \\
     & ranked first, below threshold & 0.9 & -- & -- & 1.2 & -- & 0.4 & -- & 4.7 & -- & -- & -- & -- & 0.2 & 1.1 \\
     & words fit other items & -- & -- & 27.1 & -- & 3.3 & -- & -- & -- & -- & -- & -- & -- & 5.1 & -- \\
    \addlinespace[2pt]
    Decide & match ignored & 0.3 & -- & -- & -- & 1.2 & -- & -- & -- & -- & -- & -- & -- & 0.3 & -- \\
     & match mapped to a wrong answer & 0.3 & -- & -- & -- & 1.2 & 2.1 & 2.5 & 1.9 & -- & -- & -- & -- & 0.7 & 0.7 \\
     & lure: program named a place & 8.6 & -- & -- & 5.8 & 3.3 & 5.0 & -- & -- & -- & -- & -- & -- & 2.0 & 1.8 \\
     & lure: model claimed a place & 5.8 & -- & 0.4 & 0.8 & 4.6 & 4.2 & -- & -- & -- & -- & -- & -- & 1.8 & 0.8 \\
     & stack topped out & -- & -- & -- & -- & -- & -- & -- & -- & 36.0 & 9.6 & -- & -- & 6.0 & 1.6 \\
     & caught by a ghost & -- & -- & -- & -- & -- & -- & -- & -- & -- & -- & 19.5 & 20.2 & 3.3 & 3.4 \\
     & unproductive moves & -- & -- & -- & -- & -- & -- & -- & -- & 13.1 & 6.8 & 24.8 & 20.7 & 6.3 & 4.6 \\
     & low-yield play & -- & -- & -- & -- & -- & -- & -- & -- & 3.4 & 1.5 & 6.1 & 5.0 & 1.6 & 1.1 \\
    \addlinespace[2pt]
    Execute & invalid answer & -- & -- & 3.3 & -- & -- & 0.8 & 0.4 & -- & -- & -- & -- & -- & 0.6 & 0.1 \\
     & invalid action & 2.2 & -- & -- & -- & -- & -- & -- & -- & -- & -- & -- & -- & 0.4 & -- \\
     & idle action & -- & -- & -- & -- & -- & -- & -- & -- & 9.4 & 8.8 & 10.9 & 3.3 & 3.4 & 2.0 \\
     & no action committed & -- & -- & -- & -- & -- & -- & 0.3 & -- & -- & -- & -- & -- & -- & -- \\
    \midrule
    \multicolumn{2}{@{}l}{Total points lost} & 69.8 & 0.0 & 75.8 & 13.3 & 63.8 & 37.9 & 59.6 & 26.7 & 61.9 & 26.7 & 61.3 & 49.3 & 65.4 & 25.7 \\
    \bottomrule
  \end{tabular}
\end{table*}

\section{Evolution Records}
\label{sec:supp_evo}

Table~\ref{tab:revision_records} reports completed coding sessions and changes in program size.
A session interrupted before producing a candidate may be restarted, so the number of sessions can exceed the number of candidates in Table~\ref{tab:supp_evo}.
Across the 163 candidate slots, 218 sessions were launched and 162 completed.
Two slots produced no evaluable candidate.
Interrupted sessions report no token usage, so the revision-token totals for Claude Haiku 5.5 and Qwen3.8-27B are lower bounds.

\begin{table}[t]
  \centering
  \footnotesize
  \setlength{\tabcolsep}{3pt}
  \caption{Revision sessions and program growth, summed over six tasks. Lines reports median lines added to final programs and to all candidates.}
  \label{tab:revision_records}
  \begin{tabular}{@{}lcccc@{}}
    \toprule
    & \multicolumn{2}{c}{Sessions} & \multicolumn{2}{c}{Lines added} \\
    \cmidrule(lr){2-3}\cmidrule(l){4-5}
    Method & Launched & Completed & Final & Candidates \\
    \midrule
    \multicolumn{5}{@{}l}{\textbf{GPT-6.1 Sol}} \\
    AutoMem & 30 & 30 & 500 & 536 \\
    EvoViM & 26 & 26 & 558 & 582 \\
    \multicolumn{5}{@{}l}{\textbf{Claude Haiku 5.5}} \\
    AutoMem & 44 & 33 & 674 & 624 \\
    EvoViM & 33 & 21 & 1224 & 989 \\
    \multicolumn{5}{@{}l}{\textbf{Qwen3.8-27B}} \\
    AutoMem & 41 & 30 & 833 & 1150 \\
    EvoViM & 44 & 22 & 1284 & 1144 \\
    \bottomrule
  \end{tabular}
\end{table}

The gates accept 69 of the 161 evaluated candidates.
Among the 92 rejections, 71 candidates score below the parent and 19 tie it.
The other two are AutoMem candidates on Card Matching with GPT-6.1 Sol, rejected by its malformed-request or fallback-action guard despite score gains of 5.6 and 18.1 points.
A later candidate gains 38.9 points and becomes the final program.

EvoViM accepts eight ties through its efficiency rule, six of which become final programs.
All eight parents already score 100 on the development seeds.
Below that ceiling, every accepted candidate improves the development score.
AutoMem retains the parent on all 13 ties.

For a post-hoc analysis of selection near measurement noise, we define a band using the larger of one scoring unit and the 95th percentile of score differences between two runs of the same program.
The changes for 48 rejected and 10 accepted candidates fall within this band.

\section{Control Implementations}
\label{sec:supp_controls}

\subsection{One-Shot Programs}
\label{sec:supp_oneshot}

The one-shot control starts from the same program and interfaces as EvoViM but receives no interaction records or revision experience.
References to recorded episodes are removed from its instructions, and the analysis questions remain only as design considerations.
The model can test its code on synthetic inputs, but there are no recorded observations for offline replay.
It writes one candidate without a retry, which is evaluated on the same held-out seeds as EvoViM.
Table~\ref{tab:supp_oneshot} gives the scores and answer counts.

\begin{table}[t]
  \centering
  \footnotesize
  \setlength{\tabcolsep}{3pt}
  \caption{One-shot programs and EvoViM from the same starting program. The last two rows count correct one-shot answers by query type.}
  \label{tab:supp_oneshot}
  \begin{tabular}{@{}lcccccc@{}}
    \toprule
    & \multicolumn{2}{c}{MC Gallery} & \multicolumn{2}{c}{Poster Maze} & \multicolumn{2}{c}{DMV-Hard} \\
    \cmidrule(lr){2-3}\cmidrule(lr){4-5}\cmidrule(l){6-7}
    & Haiku & Qwen & Haiku & Qwen & Haiku & Qwen \\
    \midrule
    Start & 15.0 & 23.8 & 18.8 & 23.8 & 13.8 & 7.9 \\
    One-shot & 13.8 & 13.8 & 20.0 & 25.0 & 28.8 & 14.2 \\
    \rowcolor{oursgray} EvoViM & 98.8 & 96.3 & 50.0 & 70.0 & 77.1 & 45.4 \\
    \midrule
    Seen / detail & 0 & 1 & 0 & 0 & 69 & 29 \\
    Lure / pattern & 11 & 10 & 16 & 20 & 0 & 5 \\
    \bottomrule
  \end{tabular}
\end{table}

\paragraph{Spatial tasks.}
The one-shot MC Gallery programs measure changes between consecutive frames but do not match query paintings to stored images in code.
The Claude Haiku 5.5 agent saves one frame across ten episodes.
Qwen3.8-27B saves 156 frames and recalls 105, but gives only one correct location among 60 seen queries.
On Poster Maze, both programs focus on a map of cells and routes.
Claude Haiku 5.5 stores no frame and Qwen3.8-27B stores 16 across ten episodes.
They receive 22 and 24 questions about seen posters, respectively, but neither identifies the correct wall.
Their correct answers all come from lures.

\paragraph{DMV-Hard.}
The one-shot Claude Haiku 5.5 program represents each photo with an $8\times8$ average hash and the shares of 12 color names.
Qwen3.8-27B compares color clusters and hue histograms.
These summaries of whole photos answer 69 and 29 of the 134 detail questions, but only 0 and 5 of the 106 pattern questions.

The Claude Haiku 5.5 program also introduces a mismatch between its prompt and parser.
Its record prompt requests \texttt{DONE}, while the parser accepts only command-form responses.
This causes 1,458 protocol errors and 2,139 model requests per episode.
The page records are still written, so the mismatch mainly increases request use.

\subsection{Fixed Visual Memories}
\label{sec:supp_fixed}

Each fixed visual memory replaces the starting program and runs without evolution on the same host and held-out seeds.
The execution phases and limit of 20 requests per phase remain unchanged.
We evaluate the archive and its two matcher variants on MC Gallery, Poster Maze and DMV-Hard.

\paragraph{Frame archive.}
The archive follows the interface of VISTA~\citep{han2026vista} and saves every observation losslessly by step.
\texttt{INSPECT} shows up to eight stored frames side by side, or crops the same region from each at full resolution.
\texttt{READ\_PIXELS} returns the exact colors in a region.
The text-memory files of the starting program remain available for notes.

\paragraph{Matcher variants.}
The tool variant adds \texttt{MATCH}, which searches stored frames using color zero-normalized cross-correlation.
It removes a near-uniform border from the query image or a region chosen by the model, then searches 12 template widths from 8.5\% to 79\% of the frame width.
The search tests rotations of $-15^\circ$, $0^\circ$ and $15^\circ$.
It returns the five best steps with scores and positions, and shows the query beside the three best regions.
The same routine is used for all three tasks without an acceptance threshold.

The every-step variant runs this search automatically before the first model request of each step.
It requires no additional model request and inserts the three best matches into every prompt of that step.
This variant also stores each step's text observation, such as the page URL, and presents it with the match.

\paragraph{Search scope.}
Each search covers earlier steps of the current episode, including previous question frames, and excludes the current frame.
The archive and matcher index start empty in every episode.
We choose the matcher parameters from eight variants using top-1 accuracy on the DMV-Hard development seeds, then freeze them for evaluation.
In a separate offline diagnostic on the held-out trajectories of EvoViM with Claude Haiku 5.5, this matcher ranks the correct page first for all 106 pattern queries and 85 of 134 detail queries.
This diagnostic uses the recorded observations and does not rerun the agent.

Table~\ref{tab:supp_fixed} reports both matcher variants separately.
Their paired differences from EvoViM appear in Table~\ref{tab:supp_fixed_ci}.
Table~\ref{tab:supp_fixed_cost} gives execution costs on the spatial tasks, supplementing the DMV-Hard comparison in Table~\ref{tab:dmv_cost}.

\begin{table}[t]
  \centering
  \footnotesize
  \setlength{\tabcolsep}{2.6pt}
  \caption{Fixed visual memories on the three deferred-query tasks. D / P counts correct answers among 134 photo-detail and 106 pattern queries on DMV-Hard.}
  \label{tab:supp_fixed}
  \begin{tabular}{@{}lcccc@{}}
    \toprule
    & MC & Maze & DMV & D / P \\
    \midrule
    \multicolumn{5}{@{}l}{\textbf{Claude Haiku 5.5}} \\
    \hspace{0.5em}Start program & 15.0 & 18.8 & 13.8 & 33 / 0 \\
    \midrule
    \hspace{0.5em}Archive & 20.0 & 16.3 & 22.1 & 51 / 2 \\
    \hspace{0.5em}\quad + matcher (tool) & 25.0 & 18.8 & 57.1 & 78 / 59 \\
    \hspace{0.5em}\quad + matcher (every step) & 18.8 & 16.3 & 80.8 & 98 / 96 \\
    \rowcolor{oursgray} \hspace{0.5em}EvoViM & 98.8 & 50.0 & 77.1 & 103 / 82 \\
    \multicolumn{5}{@{}l}{\textbf{Qwen3.8-27B}} \\
    \hspace{0.5em}Start program & 23.8 & 23.8 & 7.9 & 17 / 2 \\
    \midrule
    \hspace{0.5em}Archive & 25.0 & 21.3 & 20.4 & 48 / 1 \\
    \hspace{0.5em}\quad + matcher (tool) & 45.0 & 13.8 & 45.8 & 55 / 55 \\
    \hspace{0.5em}\quad + matcher (every step) & 47.5 & 15.0 & 81.7 & 98 / 98 \\
    \rowcolor{oursgray} \hspace{0.5em}EvoViM & 96.3 & 70.0 & 45.4 & 60 / 49 \\
    \bottomrule
  \end{tabular}
\end{table}

\begin{table*}[t]
  \centering
  \footnotesize
  \setlength{\tabcolsep}{8pt}
  \caption{Paired score differences between EvoViM and each fixed visual memory, with 95\% bootstrap intervals over ten seeds. Positive values favor EvoViM.}
  \label{tab:supp_fixed_ci}
  \begin{tabular}{@{}lccc@{}}
    \toprule
    & MC Gallery & Poster Maze & DMV-Hard \\
    \midrule
    \multicolumn{4}{@{}l}{\textbf{Claude Haiku 5.5}} \\
    \hspace{0.5em}Archive & 78.8 {[}72.5, 85.0{]} & 33.8 {[}20.0, 47.5{]} & 55.0 {[}49.2, 61.3{]} \\
    \hspace{0.5em}\quad + matcher (tool) & 73.8 {[}70.0, 77.5{]} & 31.3 {[}15.0, 48.8{]} & 20.0 {[}9.2, 29.6{]} \\
    \hspace{0.5em}\quad + matcher (every step) & 80.0 {[}73.8, 87.5{]} & 33.8 {[}18.8, 50.0{]} & $-$3.8 {[}$-$12.1, 4.2{]} \\
    \multicolumn{4}{@{}l}{\textbf{Qwen3.8-27B}} \\
    \hspace{0.5em}Archive & 71.3 {[}66.3, 75.0{]} & 48.8 {[}35.0, 60.0{]} & 25.0 {[}15.0, 35.0{]} \\
    \hspace{0.5em}\quad + matcher (tool) & 51.3 {[}41.3, 61.3{]} & 56.3 {[}38.8, 72.5{]} & $-$0.4 {[}$-$10.8, 10.0{]} \\
    \hspace{0.5em}\quad + matcher (every step) & 48.8 {[}36.3, 60.0{]} & 55.0 {[}37.5, 72.5{]} & $-$36.3 {[}$-$45.8, $-$27.1{]} \\
    \bottomrule
  \end{tabular}
\end{table*}

\begin{table*}[t]
  \centering
  \footnotesize
  \setlength{\tabcolsep}{3.5pt}
  \caption{Execution cost per episode on MC Gallery and Poster Maze. Req. counts model requests, In reports input tokens in millions, and Code reports seconds in program code, excluding model wait time.}
  \label{tab:supp_fixed_cost}
  \begin{tabular}{@{}lcccccc@{}}
    \toprule
    & \multicolumn{3}{c}{MC Gallery} & \multicolumn{3}{c}{Poster Maze} \\
    \cmidrule(lr){2-4}\cmidrule(l){5-7}
    Method & Req. & In & Code & Req. & In & Code \\
    \midrule
    \multicolumn{7}{@{}l}{\textbf{Claude Haiku 5.5}} \\
    \hspace{0.5em}Archive & 86 & 0.34 & 0.5 & 229 & 0.66 & 1.5 \\
    \hspace{0.5em}\quad + matcher (tool) & 88 & 0.37 & 4.1 & 248 & 0.79 & 14 \\
    \hspace{0.5em}\quad + matcher (every step) & 86 & 0.38 & 50 & 232 & 0.79 & 327 \\
    \rowcolor{oursgray} \hspace{0.5em}EvoViM & 79 & 0.36 & 37 & 416 & 2.64 & 2.3 \\
    \multicolumn{7}{@{}l}{\textbf{Qwen3.8-27B}} \\
    \hspace{0.5em}Archive & 101 & 0.29 & 1.1 & 348 & 0.73 & 5.4 \\
    \hspace{0.5em}\quad + matcher (tool) & 120 & 0.38 & 45 & 355 & 0.82 & 181 \\
    \hspace{0.5em}\quad + matcher (every step) & 118 & 0.39 & 131 & 307 & 0.80 & 1{,}560 \\
    \rowcolor{oursgray} \hspace{0.5em}EvoViM & 65 & 0.22 & 7.1 & 218 & 0.57 & 3.1 \\
    \bottomrule
  \end{tabular}
\end{table*}

\subsection{Storage in the Repeated Runs}
\label{sec:supp_repeat}

The repeated MC Gallery runs in Table~\ref{tab:supp_repeat} differ in how consistently they save observations.
The first program without revision support mentions saving only in a tool description and saves zero to six frames per episode.
The corresponding full program saves 13--18 views.
During revision, the ablated run notes that only six or seven frames were saved in each development episode, but tests its matcher as if all 16 walk frames were available.
The other two ablated runs save frames throughout the walk and reach scores of 85.0 and 78.8.

\begin{figure*}[t]
  \centering
  \includegraphics[width=\linewidth]{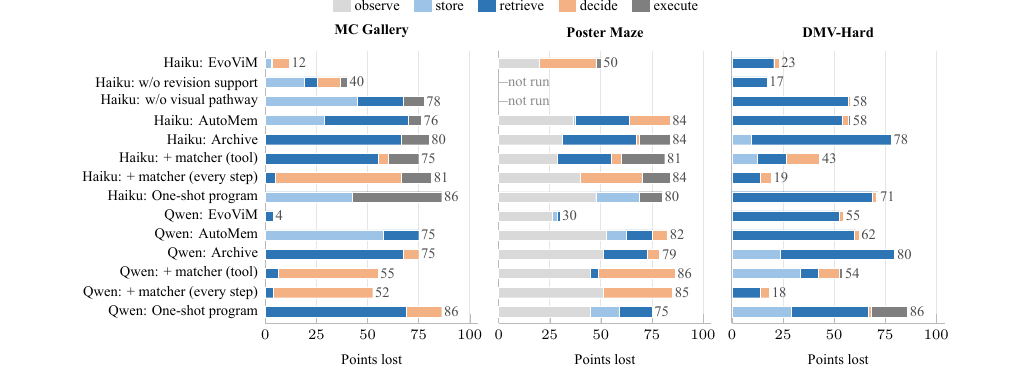}
  \caption{Score points lost at each failure stage for the ablations and fixed visual memories. For Claude Haiku 5.5 on MC Gallery, full EvoViM and the variant without revision support are averaged over three evolution runs.}
  \label{fig:fail_t3}
\end{figure*}

Figure~\ref{fig:fail_t3} applies the failure taxonomy to the ablations and fixed memories.
Across the three runs without revision support, unsaved paintings account for 17.1 lost points on MC Gallery, compared with 1.7 for full EvoViM.

\end{document}